\documentclass[amsmath, preprintnumbers, 10pt, twocolumn, pre bibliograpy, floatfix, superscriptaddress]{revtex4-1}

\usepackage{comment}
\usepackage{graphicx}
\usepackage{color}
\usepackage{svg}
\usepackage[utf8]{inputenc}

\usepackage{physics}
\usepackage{bbm}
\usepackage{braket}
\usepackage[export]{adjustbox}
\usepackage{dsfont}
\usepackage{mathrsfs}
\usepackage{multirow}
\usepackage{float}
\usepackage{tabularx}
\usepackage{caption}
\usepackage{enumitem}
\usepackage[font=small,labelfont=bf,justification=RaggedRight,format=plain, singlelinecheck=false]{caption}
\usepackage{placeins}
\usepackage{subcaption}
\usepackage{array}
\usepackage{hyperref}
\usepackage{bm}
\usepackage{cleveref}
\usepackage{amsmath}
\usepackage{amssymb}
\usepackage{stackengine}
\usepackage{mathrsfs}
\usepackage{mathtools}
\usepackage{ulem}

\renewcommand\theequation{\arabic{equation}}
\DeclareCaptionSubType*{figure}

\usepackage{xr}

\begin{document}

\title{Non-equilibrium active noise enhances generative memory in diffusion models}

\author{Agnish Kumar Behera}
\affiliation{Department of Chemistry, University of Chicago, Chicago, IL, 60637}

\author{Alexandra Lamtyugina}
\affiliation{Department of Chemistry, University of Chicago, Chicago, IL, 60637}

\author{Aditya Nandy}
\affiliation{Department of Chemistry, University of Chicago, Chicago, IL, 60637}
\affiliation{The James Franck Institute, University of Chicago, Chicago, IL, 60637}

\author{Daiki~Goto}
\affiliation{Department of Physics, University of Chicago, Chicago, IL, 60637}

\author{Carlos Floyd}
\affiliation{Department of Chemistry, University of Chicago, Chicago, IL, 60637}
\affiliation{The James Franck Institute, University of Chicago, Chicago, IL, 60637}

\author{Suriyanarayanan Vaikuntanathan}
\email{svaikunt@uchicago.edu}
\affiliation{Department of Chemistry, University of Chicago, Chicago, IL, 60637}
\affiliation{The James Franck Institute, University of Chicago, Chicago, IL, 60637}

\begin{abstract}

Generative diffusion models have emerged as powerful tools for sampling high-dimensional distributions, yet they typically rely on white gaussian noise and noise schedules to destroy and reconstruct information. Here, we demonstrate that driving the generative process out of equilibrium using active, temporally correlated noise sources fundamentally alters the information thermodynamics of the system. We show that coupling the data to an active non-Markovian bath creates a `memory effect' where high-level semantic information (such as class identity or molecular metastability) is stored in the temporal correlations of auxiliary degrees of freedom. Using Fisher information analysis, we prove that this active mechanism significantly retards the rate of information decay compared to passive Brownian motion. Crucially, this memory effect facilitates an earlier and more robust symmetry breaking (speciation) during the reverse generative process, allowing the system to resolve multi-scale structures, reminiscent of metastable states in molecular configurations that are washed out in the typical noising processes. Our results suggest that non-equilibrium protocols, inspired by active matter physics, offer a thermodynamically distinct and potentially advantageous pathway for recovering high-dimensional energy landscapes using generative diffusion. 

\end{abstract}
\maketitle

\renewcommand*{\thesection}{\Roman{section}}
\section{Introduction}
\label{Intro}
\renewcommand*{\thesection}{\arabic{section}}

Generative diffusion models are a class of machine learning models which have been used to parameterize and sample complex, high-dimensional distributions~\cite{yang2023diffusion, sohl2015deep, ho2020denoising}. Their applications range from image synthesis~\cite{Ramesh_2021_DALL-E} to scientific problems including sampling the distributions of molecular conformations~\cite{bilodeau2022generative,Wang_Herron_Tiwary_2022}, turbulent flows~\cite{whittaker2024turbulence}, and geological modeling~\cite{Lochner_2023_terrain_diffusion}. In these models, samples from the training dataset are first transformed into multidimensional Gaussian distributions (with variance specified by the hyperparameters of the model) through a process analogous to overdamped Brownian diffusion in a harmonic potential~\cite{sohl2015deep, anderson1982reverse}. In score-based diffusion models, during the ``forward'' phase, a neural network (NN) is trained to learn the score function of the distribution, which encodes information about how the data samples are progressively transformed into Gaussian white noise. This process, based on standard stochastic calculus techniques with inspiration from non-equilibrium thermodynamics~\cite{sohl2015deep}, is very effective at parameterizing the unknown target distribution from which the training samples are drawn. Combined with machine learning architectures such as U-nets, this approach can produce new samples from a high-dimensional target distributions (e.g., images) that are strikingly similar to the original data~\cite{biroli2023generative,biroli2024dynamical}. While much effort has focused on improving neural network architectures and training procedures~\cite{kingma2021variational, karras2022elucidating, cao2023exploring, haas2024discovering, tzen2019theoretical, chen2023speed, phung2023wavelet}, a fundamental question remains largely unexplored: \textit{what physical diffusion process is optimal for learning and sampling?}  Specifically, although the widely successful current score-based diffusion models~\cite{Song_2021_SGM_SDE} rely on a forward process governed by effectively, \textit{passive} diffusion with uncorrelated Gaussian white noise~\cite{sohl2015deep, anderson1982reverse}, there is little reason \textit{a priori} to expect that the simplistic setting of overdamped Brownian dynamics provides in all cases the optimal physical model on which to base the diffusion process. Here, inspired by active matter physics, where particles exhibit persistent, correlated motion~\cite{fodor2016far,Seifert_2012}, we ask: can generative diffusion with non-equilibrium dynamics akin to active matter, referred to as active diffusion below, help improve performance of these generative models?



We systematically validate the generative capabilities of this non-equilibrium framework across a hierarchy of complexity, ranging from low-dimensional toy models and molecular conformations to the model high-dimensional datasets. In scenarios governed by complex, multi-scale geometries we find that active diffusion significantly outperforms standard passive dynamics, faithfully reconstructing features that are otherwise washed out by uncorrelated noise (Fig.~\ref{fig:schematic},Fig.~\ref{fig:complex_toy_dsns:multimodal_swissroll_overlap}). To identify the physical mechanism driving this enhancement, we employ information-theoretic tools to quantify the system's memory retention, revealing that active non-equilibrium dynamics possess a fundamentally slower rate of information decay~\cite{Ganguli-Huh-Sompolinsky08} (Fig.\ref{fig:FMC}). We demonstrate the practical consequence of this extended memory by examining the stability of categorical information~\cite{sclocchi2025phase} within the MNIST landscape (Fig.~\ref{fig:complex_toy_dsns:mnist}), observing that active noise preserves the distinct structural features that define a specific digit significantly longer than passive methods. Crucially, we show that this robustness arises because the correlated auxiliary variables (Fig.~\ref{fig:schematic}) inherent to the active process do not merely act as noise; rather, they actively store the categorical identity of the data (Fig.~\ref{fig:lostcorrelations}). These results are supplemented by analytical characterization of a hierarchical data model~\cite{sclocchi2025phase} where we show how the active noise is able to guide recovery of categorical identity even when the noised data has completely lost this information. This again demonstrates how the non-equilibrium correlations due to the active dynamics can have beneficial generative consequences.
We postulate that this feature effectively partitions the generative task: by offloading the maintenance of global class identity to the auxiliary active variables, the generative dynamics are liberated to focus their capacity on resolving fine-scale, local fluctuations. Consequently, rather than expending the reverse trajectory on rediscovering which image class to generate, the model can dedicate the diffusion process to refining the intricate microscopic realizations of that digit—a strictly advantageous regime that manifests as sharper resolution in multi-scale distributions and clearer separation of metastable molecular state. Finally, and consistent with the aforementioned findings, we show how the speciation times corresponding to active generative processes is sooner, allowing the generative process to focus on the finer structure. We finally note that correlations between data and active noise degrees of freedom when viewed through the lens of active matter leads to terms like active pressure or dissipation~\cite{fodor2016far}. Our work shows how these same correlations can help with data generation in generative diffusion.


This paper is organized as follows. In Sec.~\ref{sec:ActiveRevTimeDif}, we introduce the analytical theory of reverse-time diffusion in the presence of active noise-assisted forward process. In Sec.~\ref{sec:Experiments} we present the various datasets to which we apply our novel diffusion scheme and analyze its performance. Finally, in Sec.~\ref{sec:ClassStructure} and Sec.~\ref{sec:Discussion} we propose possible mechanisms through which the correlated noise sources might be helping improve the generative properties of the diffusion process.

\renewcommand*{\thesection}{\Roman{section}}
\section{Reverse-time diffusion in the presence of active noise}
\label{sec:ActiveRevTimeDif}
\renewcommand*{\thesection}{\arabic{section}}

\begin{figure*}
    \centering
    \includegraphics[width=0.9\linewidth]{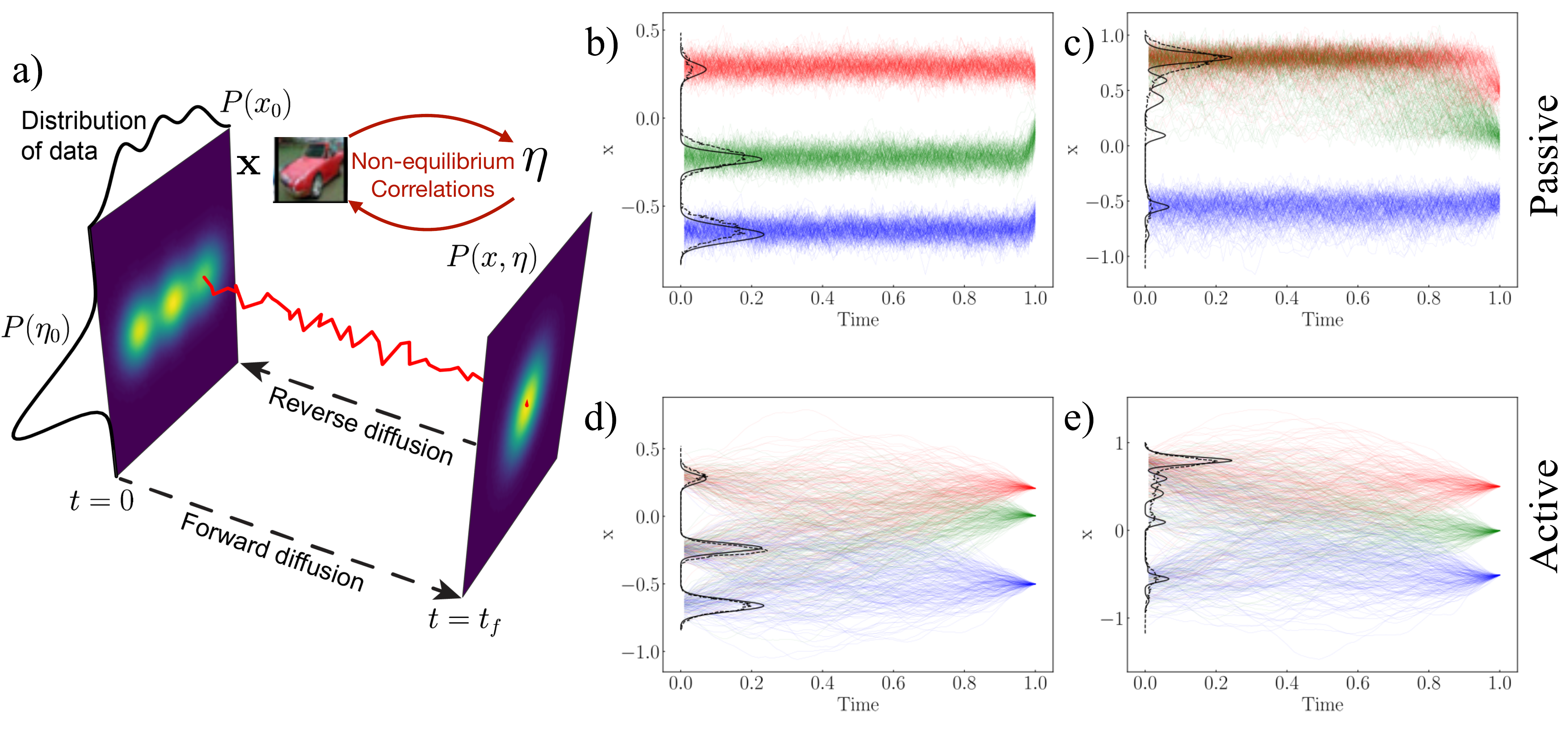}
    \caption{Schematic for the forward and backward diffusion processes. (a) Active diffusion correlates a noise variable $\eta$ with the data degrees of freedom ${\bf x}$  during generative diffusion processes. (b), (c), (d) and (e) denote passive and active reverse diffusion for two one dimensional distributions: (b,d) is a \textit{coarse}r distribution having three almost non-overlapping distinct peaks with all three of them having an almost equal weight whereas (c,e) his a \textit{finer} distribution having five closely overlapping peaks with 4 peaks having a much smaller weight than the one large peak. In the case of distributions in (c, e) the active process better resolves finer scale features.}
    \label{fig:schematic}
\end{figure*}

We first review one of the standard generative diffusion frameworks. We will refer to this as ``passive" diffusion~\cite{biroli2023generative}, and it is equivalent to the diffusion process described in Ref.~\cite{Song_2021_SGM_SDE}. In the passive forward process a given data distribution is evolved according to the following equation of motion:
\begin{equation}
  \dot{\textbf{x}} = -k\textbf{x} + \bm{\xi}(t) \label{eq:PassiveForward}   \ ,
\end{equation}
where $\textbf{x}$ is the $d$-dimensional data point and the Gaussian noise $\bm{\xi}(t)$ has the properties $\langle \xi_i(t) \rangle = 0$ and $\langle \xi_i(t) \xi_j(t') \rangle = 2T\delta_{ij} \delta(t-t')$ for all $i,j \in \{1,2,\ldots,d\}$, with $\langle \cdot \rangle$ denoting an ensemble average over independent noise realizations. 
This evolution systematically destroys the correlations in the data. The temperature $T$ and stiffness $k$ are hyperparameters that set the timescale of relaxation and the width of the multidimensional isotropic Gaussian distribution, with mean $0$ and variance $T/k$, that the system eventually settles into. 
The reverse diffusion process then reconstructs the data distribution back from the Gaussian distribution, with dynamics given by
\begin{equation}
  -\dot{\textbf{x}} = -\textbf{x} + 2T\mathscr{F}(\textbf{x};t) + \bm{\xi}(t) \ ,
\end{equation}
where $\mathscr{F}(\textbf{x};t) \equiv \nabla_{\textbf{x}} \log{P(\textbf{x},t)}$ is the score function that helps guide the reverse trajectories to the original distribution~\cite{anderson1982reverse}. The exact form of the score function $\mathscr{F}(\textbf{x},t)$ depends on the initial distribution from which the data is drawn. For most distributions, the score function cannot be calculated analytically and is instead approximated from data using neural network models. The loss function $\mathscr{L}$ guiding the construction of the neural network model is given as a mean-squared error between the true score function and that calculated from training data. Following~\cite{biroli2023generative}, an expression for $\mathscr{L}$ can be derived using
\begin{equation}
    \frac{\partial \log{P({\bf x},t)}}{\partial x_i} = -\frac{x_i - \braket{x_{0,i}}_{P(\mathbf{x}_0|\mathbf{x},t)}e^{-kt}}{\Delta_t} \ ,
\end{equation}
where $\Delta_t \equiv \frac{T}{k}(1-e^{-2kt})$, $\textbf{x}_0$ is the data configuration at $t=0$, and $\langle{\cdot}\rangle_{P(\mathbf{x}_0|\mathbf{x},t)}$ denotes an average with respect to the conditional (posterior) distribution $P(\mathbf{x}_0|\mathbf{x},t)$.
Using this, the mean-squared-error loss function is computed as
\begin{align}
    \mathscr{L} =& \int d\textbf{x} \, P(\textbf{x},t) \lVert S_{\textbf{w}}(\textbf{x}) - \mathscr{F}(\textbf{x},t) \rVert^2 \ .
\end{align}
Here, $S_{\textbf{w}}(\textbf{x})$ is the neural network model for the score function, and $\textbf{w}$ denotes the weights of the neural network that are optimized using stochastic gradient descent algorithms.

Building on this existing framework, we now describe our ``active" generative diffusion process. Under the influence of active noise, the forward process is
\begin{align}
  \dot{\textbf{x}} &= -k\textbf{x} + \bm{\eta}(t) + \bm{\xi}_1(t) \label{eq:Activex} \\
  \dot{\bm{\eta}} &= -\frac{\bm{\eta}}{\tau} + \bm{\xi}_2(t) \label{eq:ActiveEta} \\
  \langle \xi_{1,i}(t) \rangle &= 0 , \quad \langle \xi_{2,i}(t) \rangle = 0 ,\\
  \langle \xi_{1,i}(t) \xi_{1,j}(t') \rangle &= 2T_p\delta_{ij} \delta(t-t') \, \forall \, i,j \\
  \langle \xi_{2,i}(t) \xi_{2,j}(t') \rangle &= \frac{2T_a}{\tau^2} \delta_{ij} \delta(t-t') \, \forall \, i,j \ .
\end{align}
As shown in the schematic of Fig.~\ref{fig:schematic}, every ``data" degree of freedom, $\textbf{x}$, has an ``active" degree of freedom, $\bm{\eta}$ associated with its evolution. In essence the dimension of the system is increased from $d$ to $2d$, where $d$ is the dimension of the data. In Sec.~\ref{appsec:Derivation} we show that the reverse diffusion for this process is given by
\begin{align}
  -\dot{\textbf{x}} =& -k\textbf{x} + \bm{\eta} + 2T_p \mathscr{F}_\textbf{x}(\textbf{x},\bm{\eta};t) + \bm{\xi}_1(t) \\
  -\dot{\bm{\eta}} =& -\frac{\bm{\eta}}{\tau} + \frac{2T_a}{\tau^2} \mathscr{F}_{\bm{\eta}}(\textbf{x},\bm{\eta};t) + \bm{\xi}_2(t) \ ,
\end{align}
where $\mathscr{F}_{\textbf{x}}(\textbf{x},\bm{\eta};t)\equiv  \nabla_{\bm{x}} \log P(\textbf{x},\bm{\eta};t)$ and $\mathscr{F}_{\bm{\eta}}(\textbf{x},\bm{\eta};t)\equiv  \nabla_{\bm{\eta}} \log P(\textbf{x},\bm{\eta};t)$ are the score functions for this process. 

As in the passive case, one can construct the loss function for training the neural network as
\begin{align}
    \mathscr{L} =& \int d\textbf{x} d\bm{\eta} \, P(\textbf{x},\bm{\eta};t) \bigg[ \lVert S^{(\textbf{x})}_\textbf{w}(\textbf{x},\bm{\eta}) - \mathscr{F}_\textbf{x}(\textbf{x},\bm{\eta};t)  \rVert^2 \nonumber \\
    &+ \lVert S^{(\bm{\eta})}_\textbf{w}(\textbf{x},\bm{\eta}) - \mathscr{F}_{\bm{\eta}}(\textbf{x},\bm{\eta};t)  \rVert^2 \bigg] \ .
\end{align}
Here $S_{\textbf{w}}^{(\textbf{x})}$ and $S_{\textbf{w}}^{(\bm{\eta})}$ are two different neural networks used for approximating the score in $\textbf{x}$ and the score in $\bm{\eta}$, respectively. We derive the forms of $\mathscr{F}_\textbf{x}(\textbf{x},\bm{\eta};t)$ and $\mathscr{F}_{\bm{\eta}}(\textbf{x},\bm{\eta};t)$ in Sec.~\ref{appsec:Derivation}. This choice leads to training of only the neural network for $\mathscr{F}_{\bm{\eta}}(\textbf{x},\bm{\eta}; t)$ since $\mathscr{F}_{\textbf{x}}(\textbf{x},\bm{\eta}; t)$ becomes irrelevant for the reverse process. 

In Fig.~\ref{fig:schematic} (b-d), we qualitatively show how active and passive generative dynamics can start to differ for a simple one dimensional landscape. In Fig.~\ref{fig:schematic} (b, d) we compare the trajectories seen in the two processes as they seek to recreate a distribution with three modes. In the passive case (b), trajectories commit to one of the modes early and then finer features are sampled. In the corresponding active case (d), the reverse process is more ergodic allowing the landscape to be better sampled. The benefits of such sampling are more dramatic in (c, e) where rare modes in the distribution are sampled much more efficiently in the active case. We note that such improvements in the sampling effectiveness are in line with theoretical work on the diffusion of active particles in rugged landscapes~\cite{Dor2019}. This result, on a minimal one dimensional setup, shows the potential promise of using active processes to enable generative diffusion. We also note that this qualitative result cannot be used to immediately comment on the so called speciation times in generative diffusion processes. We focus on that explicitly in the later part of the manuscript.   
We also note that a similar approach of expanding the dimensionality through additional degrees of freedom is taken in \cite{dockhorn2021score}, which presents an underdamped passive Brownian diffusion process (referred to as critically-damped Langevin diffusion, CLD).  In that case, destructive noise is not added to the data directly but instead to the degrees of freedom. Note however, that our method can access non-equilibrium regimes that are not allowed in the CLD method and hence these two are not equivalent. Indeed, in many cases (as reported in the SI), we find that the numerical performance of our method is better than the CLD method. 
Implementation details are discussed in Sec.~\ref{appsec:NumericalImplementation}. In the next section, we discuss various numerical experiments where we compare the performance of passive and active generative diffusion models. 

\renewcommand*{\thesection}{\Roman{section}}
\section{Reverse generative dynamics with active score functions}
\label{sec:Experiments}
\renewcommand*{\thesection}{\arabic{section}}

\begin{figure*}[ht]
    \begin{subfigure}[t]{0.45\textwidth}
    \centering
    \caption{(a) Analytic score\label{fig:multigaussian:analytical}}
    \includegraphics[width=0.9\linewidth]{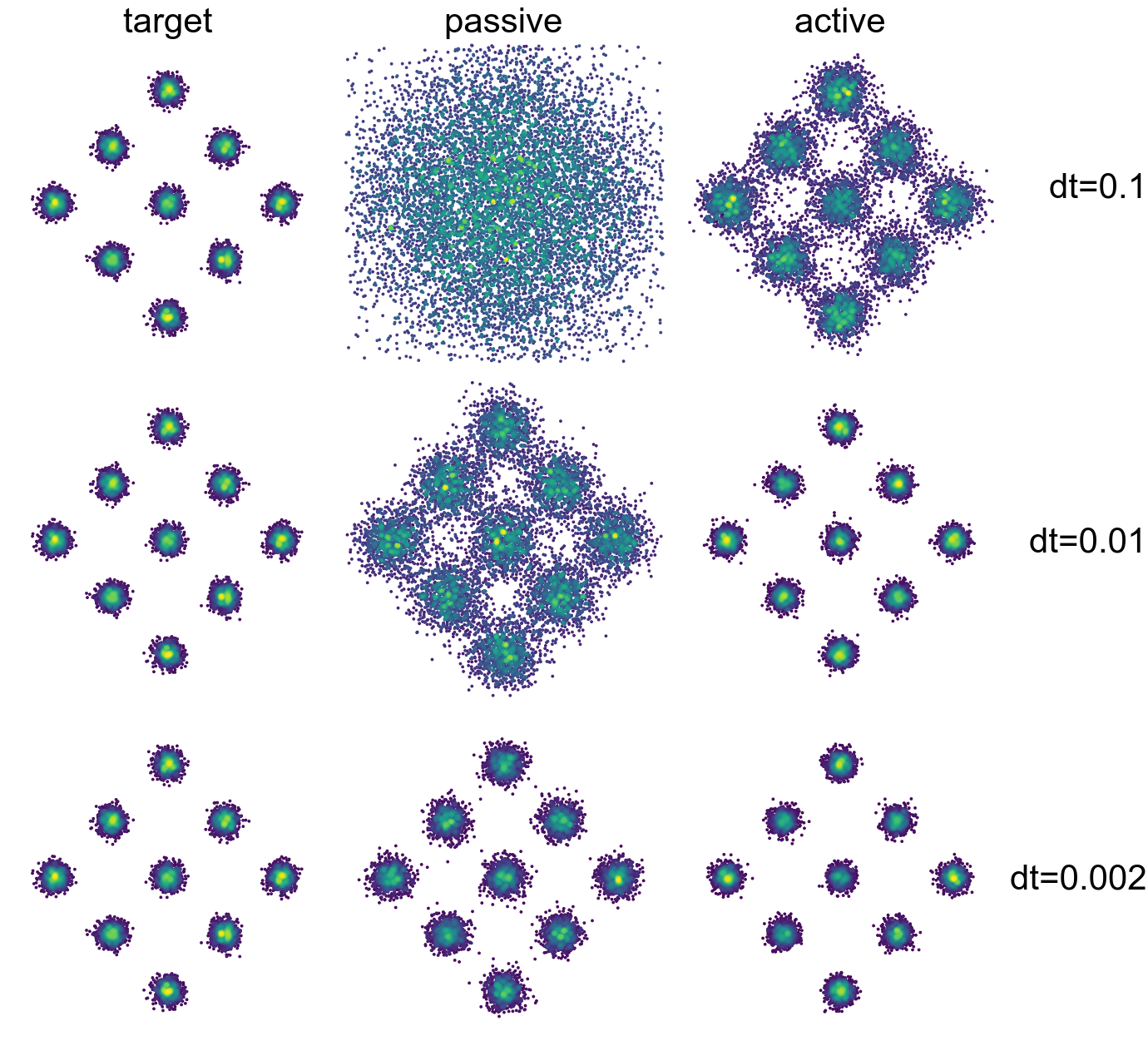}
    \end{subfigure}%
~
    \begin{subfigure}[t]{0.45\textwidth}
    \centering
    \caption{(b) Neural network score\label{fig:multigaussian:NN}}
    \includegraphics[width=0.9\linewidth]{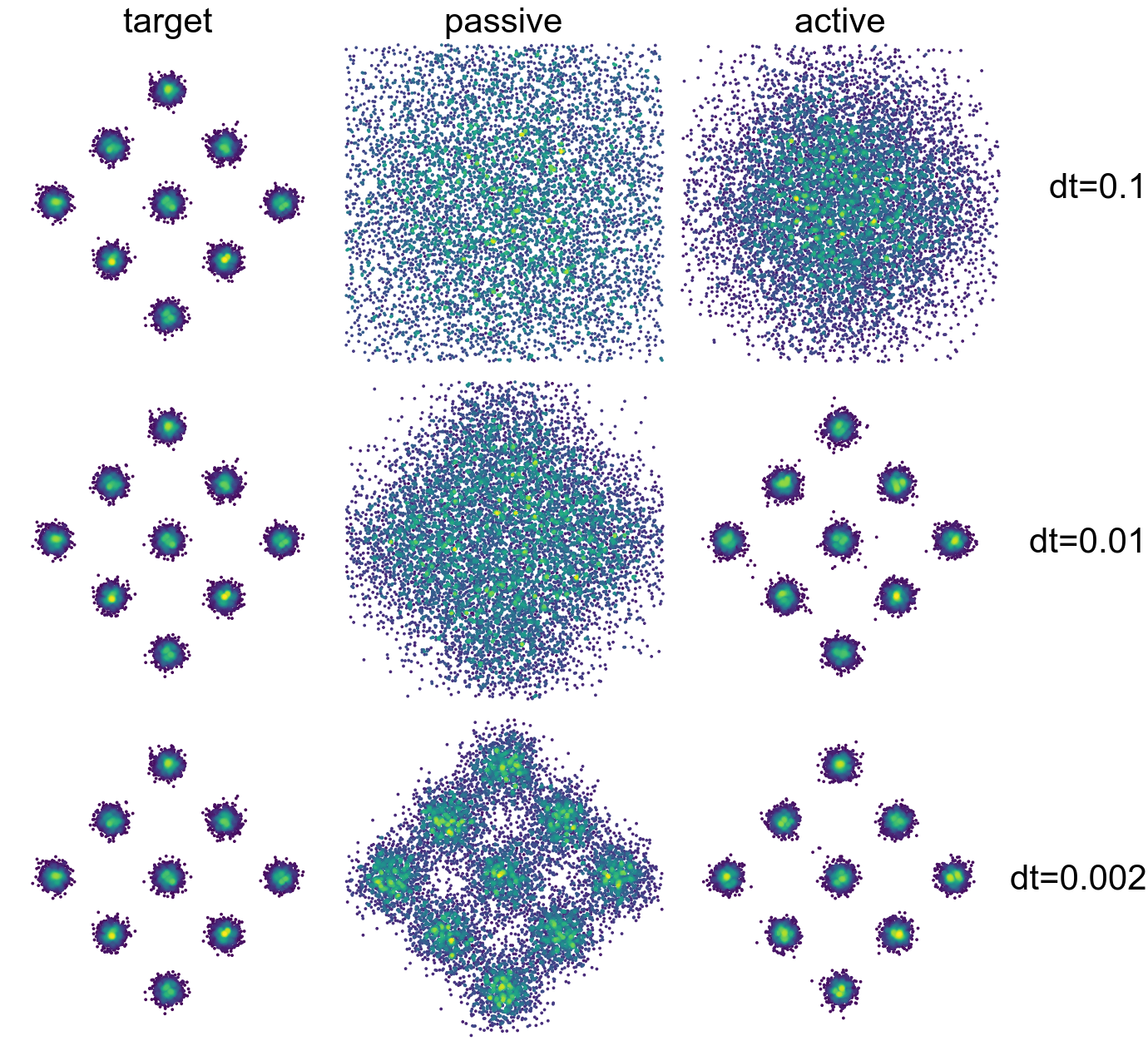}
    \end{subfigure}%
~
    \caption{Generation of samples in a Gaussian mixture distribution via reverse diffusion with (a) analytical scores and (b) scores learned with a neural network from training data. Each scatter plot contains 10,000 2D samples, colored by sample density for visual clarity (higher densities are indicated by green points and lower densities by purple). (a) Active diffusion outperforms passive case for larger $dt$, the time step size of the reverse diffusion trajectory. As $dt$ decreases, the performance of passive diffusion becomes comparable to that of active diffusion. For very small $dt$, both passive and active diffusion accurately reproduce the target distribution. (b) When using neural networks to approximate score function in the reverse diffusion process, passive and active diffusion show similar trends as those seen in diffusion with the analytic score function. Both improve as $dt$ is decreased, but the overall performance is worse than in the analytic example. The step size \textit{dt} determines the number of sampling steps in the reverse-diffusion process. Smaller \textit{dt} imply a larger number of timesteps which incur a huge overhead cost in terms of calling the neural network function. \label{fig:multigaussian} }
\end{figure*}

The performance of score-based generative diffusion is governed by the ability to accurately learn the score function and efficiently evolve the equations of motion of the reverse diffusion process. 
In this section, we examine the effects of approximating the score function on the performance of passive and active diffusion. First, we examine diffusion performance on a target distribution for which the analytic form of the score is known.  Then, we numerically approximate the score function with a multi-layer perceptron (MLP) from data and examine the effects of discretization of the learned score function in the sampling process of reverse diffusion. Details on the neural network architectures and numerical implementation are in Sec.~\ref{appsec:NumericalImplementation}.

\subsection{Gaussian mixture 2D distribution: analytic score vs. score modeled by a neural network}
\label{sec:analytic_score}

The score function for a simple distribution such as a mixture of Gaussian peaks can be expressed analytically. The analytical score functions for such distributions can be used to compare the performance of passive and active diffusion without needing to account for the learning performance of neural networks, since in this case the score functions are known exactly and do not need to be inferred from training data. In these numerical experiments, we investigate the performance with respect to the time step size for the reverse diffusion process. The time step step size is given by $dt=t_f/n$, where $t_f$ is the total time of a trajectory and $n$ is the number of sampling steps. For all diffusion trajectories presented here, $t_f=1$.

The Gaussian mixture distribution is a typical simple distribution to test the performance of the neural networks in generating the reverse diffusion process. The score functions of Gaussian distributions have an exact analytic form. The data distribution we generate is given by
\begin{align}
    P_0(\textbf{x}_0) \propto & \sum_{\alpha} \frac{p_{\alpha}}{\Pi_{i}\sqrt{h_{i}^{\alpha}}} \exp \left(-\sum_{i} \frac{(x_{0,i} - \mu_i^{\alpha})^2}{2h_i^{\alpha}} \right) \ ,
\end{align}
where $i$ denotes the dimension of the data, $\mu_i^{\alpha}$ and $h_i^{\alpha}$ describe the location of the mean and the corresponding variance, respectively, of the $\alpha^{th}$ Gaussian in $i^{th}$ dimension, and $p_{\alpha}$ represents the weight given to each of the Gaussian peaks in the mixture. The details of the derivation of the score functions for the passive and active processes are provided in Sec.~\ref{appsec:toy_models}.

We use a distribution where 9 Gaussians are spaced in a diamond formation (see Sec.~\ref{appsec:toy_models} for parameters used to generate the distributions). We perform reverse diffusion using the analytical score function and the score function learned by a neural network (Fig.~\ref{fig:multigaussian}).

Using the analytical score function for the reverse process, we show in Fig.~\ref{fig:multigaussian:analytical} that for large time step sizes ($dt=0.1$ and $dt=0.01$), the active process outperforms its passive counterpart. If the time step is small enough ($dt=0.002$), both passive and active diffusion achieve comparable performance, and both are capable of faithfully reproducing the target distribution.  When we use neural networks to learn the score function numerically for the same data distribution (Fig.~\ref{fig:multigaussian:NN}), passive and active processes show similar trends as in the analytical case in Fig.~\ref{fig:multigaussian:analytical}, although the overall performance is lower than when the analytical score is used. We note that here, for exact comparison between the analytical case and the case with neural networks, we turn off denoising in the final step of the reverse diffusion process. Generally, for passive diffusion processes, the last step of reverse diffusion is carried out with only the drift term and the noise term is set to zero. This last denoising step has been observed to improve FID scores in image datasets~\cite{Jolicoeur-Martineau_2021}. In our case of active reverse diffusion, the denoising step has no effect as the denoising is applied to the $\bm{\eta}$ dimension. However, in passive reverse diffusion, the denoising step affects the quality of the generated data since the denoising is applied to the data directly. Additional details are provided in Sec.~\ref{appsubsec:sampling}.

\subsection{Distributions with unknown score functions}

\begin{figure}[htb]
\begin{subfigure}[t]{0.4\textwidth}
    \centering
    \caption{(a) Overlapping Swiss rolls}
    \includegraphics[width=\linewidth]{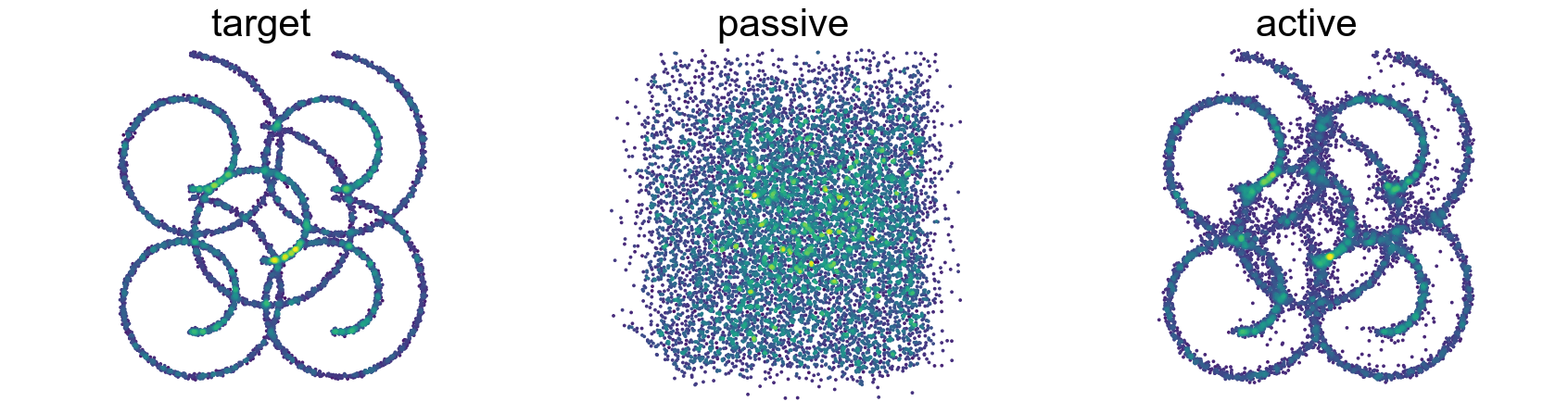}
    \label{fig:complex_toy_dsns:multimodal_swissroll_overlap}
    \end{subfigure}
~
    \begin{subfigure}[t]{0.4\textwidth}
    \centering
    \caption{(b) Alanine dipeptide $(\phi,\psi)$, 2D diffusion}
    \includegraphics[width=0.95\linewidth]{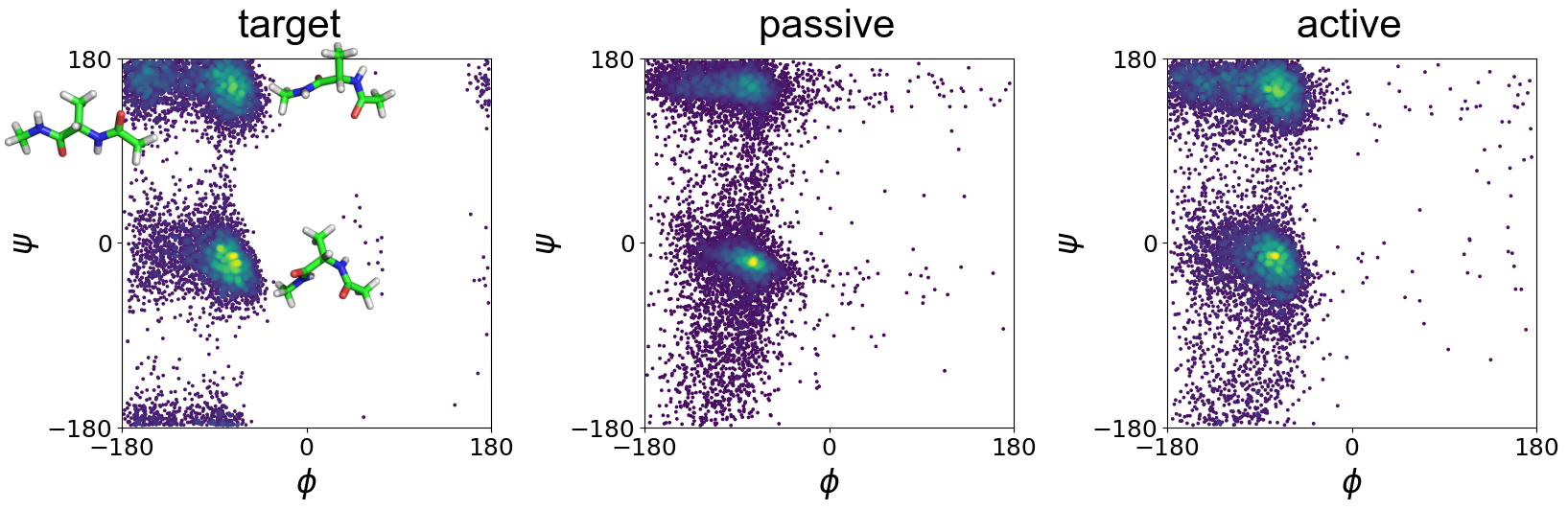}
    \label{fig:complex_toy_dsns:alanine_dipeptide}
    \end{subfigure}
    \vskip\baselineskip
~
    \begin{subfigure}[t]{0.4\textwidth}
    \centering
    \caption{(c) Alanine dipeptide (first two principal components), 25D diffusion}
    \includegraphics[width=0.95\linewidth]{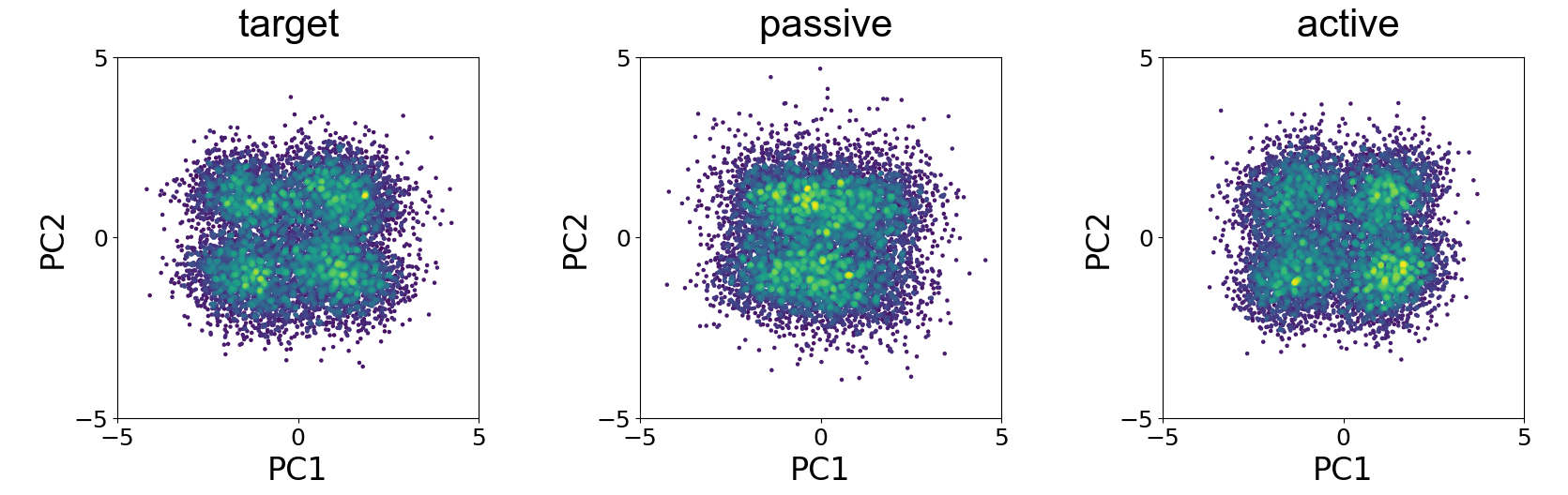}
    \label{fig:complex_toy_dsns:alanine_dipeptide_25}
    \end{subfigure}
    \vskip\baselineskip

    
~
    \caption{(a) Generation of samples for a 2D distribution of multiple overlapping Swiss rolls via reverse diffusion with the score function approximated by a neural network. (b) Ramachandran plots ($\phi$, $\psi$) in degrees for 1 $\mu$s of molecular dynamics sampling for a water-solvated alanine dipeptide (left) and corresponding diffusion generated samples with passive (center), and active ($\tau=0.5$) (right). The score model was trained using a MLP on a 2D dataset consisting of dihedral angle pairs $(\phi, \psi)$. (c) The first two principal components of alanine dipeptide configuration parameters. The score model was trained using a U-net on a 25D dataset consisting of bond lengths, bond angles, and dihedral angles.
    }
    \label{fig:complex_toy_dsns}
\end{figure}

\begin{figure}
    \centering
    \includegraphics[width=0.8\linewidth]{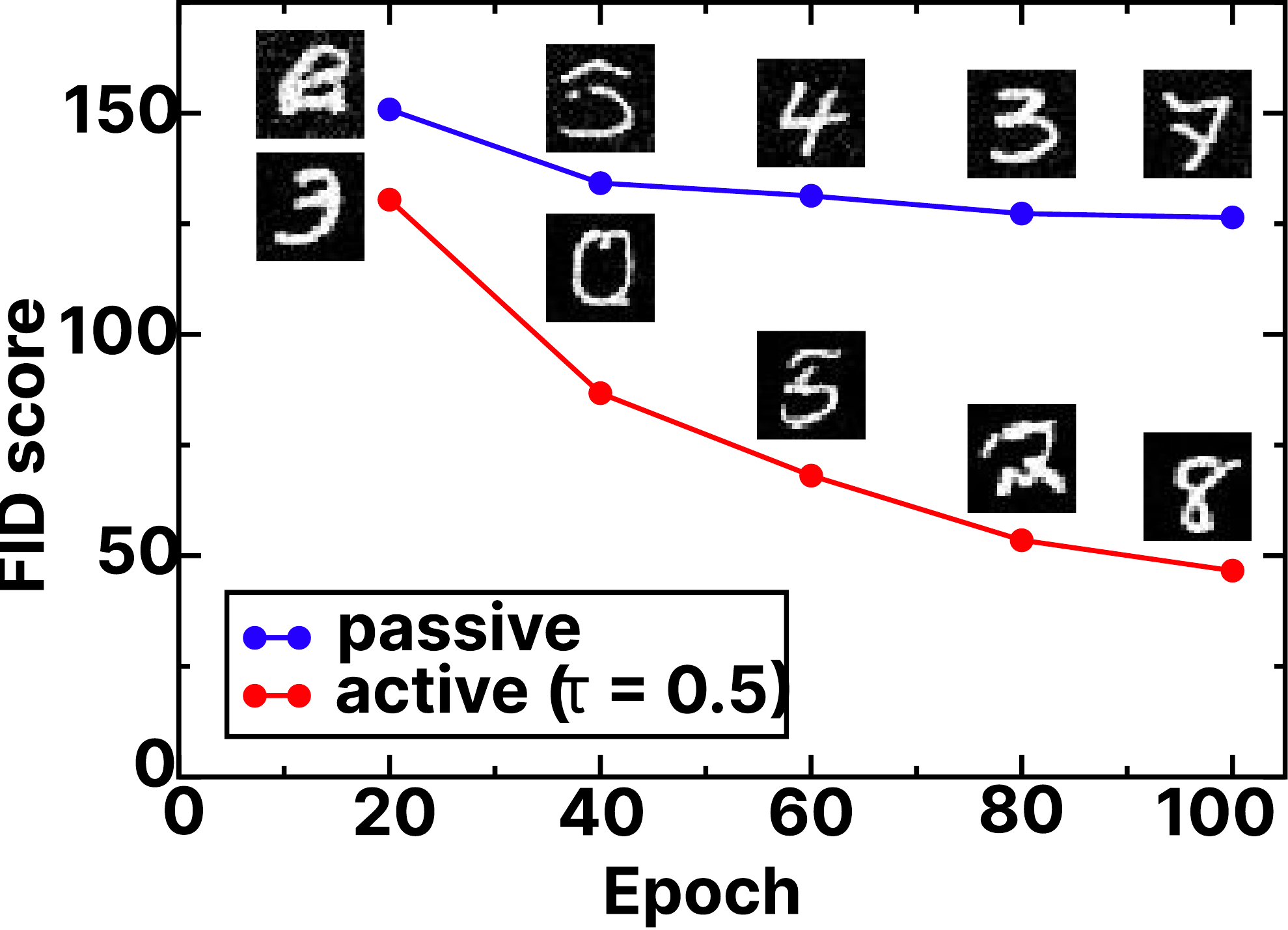}
    \caption{Fréchet Inception Distance (FID) scores as a function of training epoch number for passive (blue) and active (red) diffusion models with $\tau = 0.5$. Representative generated digits are shown for models trained at every 20 epochs.}
    \label{fig:complex_toy_dsns:mnist}
\end{figure}

We next test the performance of passive and active diffusion on 2D distributions with reduced symmetry and increased multi-scale structure than the diamond of Gaussians. To compare passive and active processes, we consider a data distribution consisting of overlapping Swiss rolls (Fig.~\ref{fig:complex_toy_dsns:multimodal_swissroll_overlap}) to test the method on a distribution for which the analytical form of the score function is not known. An important feature of this example is that the true distribution has structure at multiple length scales: both the position of the rolls and their interior structures need to be captured by the generative diffusion model.  We use neural networks to learn the score functions, and, as in the Gaussian mixture model example, we observe that active diffusion outperforms passive diffusion (Fig.~\ref{fig:multigaussian:NN}). Additional results for this and other toy models are included in Sec.~\ref{appsec:toy_models}.  We observe that the passive process is unable to generate the target distribution for all values of time step size and for the entire range of iterations that we have considered in Fig.~\ref{si:fig:multimodal_swissroll_overlap}, while the active process succeeds in accurately resolving both scales of the target distribution.   

We also consider the alanine dipeptide molecule, a small model benchmark system whose fluctuations can be numerically simulated for long timescales to generate training data (Fig.~\ref{fig:complex_toy_dsns:alanine_dipeptide}). We started with the geometry of the alanine dipeptide from a previous benchmarking study~\cite{Nuske2017Alanine}. The details of the training data generation procedure are given in Sec.~\ref{appssec:ADTrainingData}. For this sampling, we computed the Ramachandran dihedral angles ($\phi$, $\psi$) for all conformations. Based on the energy landscape of the alanine dipeptide, three major conformations~\cite{Weise2003Alanine} ($\alpha_R$ [($\phi$, $\psi$) = (-60$^\circ$, -45$^\circ$)], $P_{II}$ [($\phi$, $\psi$) = (-75$^\circ$, 145$^\circ$)], $C_5$ [($\phi$, $\psi$) = (-180$^\circ$, 180$^\circ$)]) emerge from the molecular dynamics simulation, with other conformations sampled less frequently. We use generative diffusion to resample the two-dimensional ($\phi$, $\psi$) landscape. Relative to passive diffusion, active diffusion is able to better reproduce the conformational landscape sampled during molecular dynamics. In particular, the distribution of the Ramachandran dihedral angles of the $C_5$ conformation are better reproduced by active diffusion relative to passive diffusion. While the positions of the $\alpha_R$ and $P_{II}$ angle distributions are also reproduced by both models, the separation between them is more evident in the samples generated by the active model in the same number of iterations.

We also perform diffusion on the full 25-dimensional dataset consisting of all of the parameters required to describe the conformation of an alanine dipeptide molecule (Fig.~\ref{fig:complex_toy_dsns:alanine_dipeptide_25}). The score model for the diffusion of the 25D dataset was approximated with a U-net. To compare the effectiveness of the different types of diffusion to the training data, we visualize the first two principal components of the datasets.

As in the case of the lower-dimensional toy datasets, active diffusion better reproduces the target distribution at a lower number of training iterations than passive diffusion. Fig.~\ref{si:fig:ala_25} demonstrates the improvement in sample quality as the number of training iterations is increased.

Finally, we consider the MNIST dataset as an example of a well-studied, high-dimensional dataset. The MNIST dataset\cite{lecun_2018_gradientbased} is a widely used benchmark in machine learning, consisting of 70,000 grayscale 28x28 pixel images of handwritten digits (0-9) appearing as white pixels on a black background. The full dataset is partitioned into a standardized 60,000/10,000 train/test split. Although MNIST is simple compared to modern high-resolution image datasets such as ImageNet\cite{deng_2009_imagenet}, it remains valuable for benchmarking due to its manageable size and well-defined evaluation metrics. These characteristics make it particularly suitable for investigating fundamental properties of diffusion models.

We trained both passive and active diffusion models using on the MNIST training set with the same underlying U-Net architecture and hyperparameters. To quantify the quality of samples, the Fréchet Inception Distance (FID) score\cite{heusel_2017_gans} has emerged as the \textit{de facto} standard for evaluating the quality of images produced by generative models. The FID score measures the statistical similarity between the distribution of real and generated images by comparing their activations in the feature space of a pre-trained neural network. FID scores are particularly valuable because they capture both the fidelity of individual generated samples and the diversity of the overall distribution. Lower FID scores indicate generated distributions that more closely match the real data distribution. Following standard practice, we employed the Inception-V3 network\cite{szegedy_2015_rethinking} pre-trained on ImageNet as the feature extractor for FID score calculations. While originally designed for RGB images at higher resolutions, we adapted the MNIST grayscale images by replicating the single channel across three channels and resizing from 28×28 to 299×299 pixels using bilinear interpolation before feeding them into the network. Features were extracted from the final pooling layer (2048-dimensional feature vectors), providing a rich representation space for comparing the statistical properties of image distributions. All FID scores were computed using the PyTorch implementation of FID.\cite{maximilianseitzer_2021_pytorchfid}

Computing the FID scores over training epoch number revealed notable differences between passive and active diffusion models. Across all training durations, active diffusion models consistently achieved lower (i.e. better) FID scores compared to the passive counterpart (Fig.~\ref{fig:complex_toy_dsns:mnist}). This difference was particularly pronounced in the early stages of training, suggesting that active diffusion accelerates the convergence to generating high-quality samples. Visual inspection of the generated samples revealed that a significant portion of the FID score improvement can be attributed to background quality of the MNIST digits. Samples from passive diffusion models frequently exhibited spurious white and gray specks in the background areas, particularly during early training stages. These artifacts gradually diminished with extended training but persisted even after 100 epochs. In contrast, active diffusion models consistently produced solid black backgrounds from relatively early in the training process, leading to cleaner sample generation and consequently better FID scores. 
\begin{figure}[ht]
    \centering
    \includegraphics[width=0.95\linewidth]{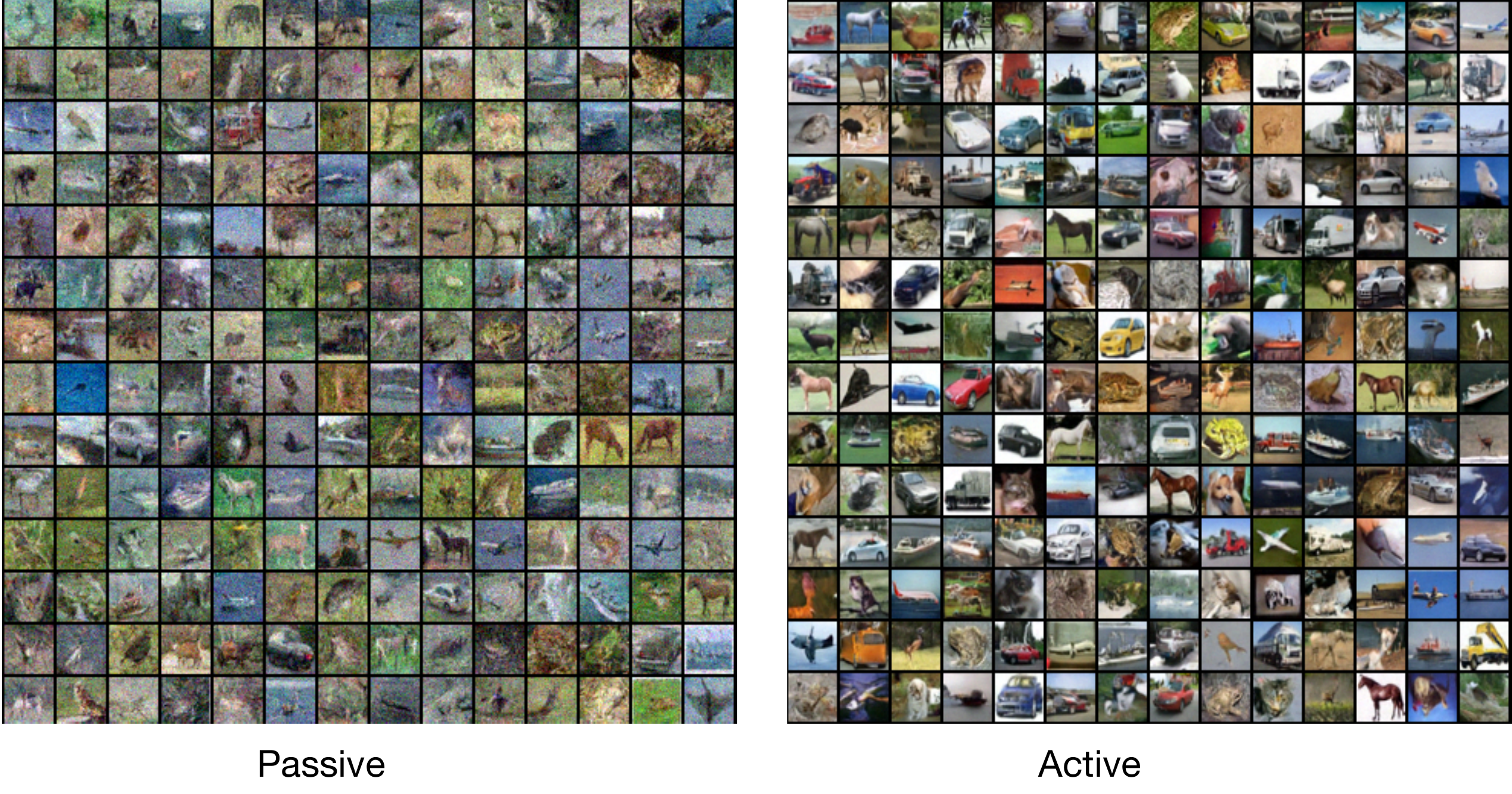}
    \caption{Comparing generated images with active and passive versions of generative diffusion with the CIFAR-10 dataset. Parameters used $k=4$, $T_a=6.4$, $\tau=0.15$ for active and similar $k,T_p$ parameters for passive \label{fig:cifar}. A total of $\sim  \times 10^5$ steps were used in both cases. The active version performs better than its passive counterpart in line with previous findings. }
\end{figure}

Finally, we trained both active and passive versions of the diffusion model on a CIFAR-10 dataset with parameters $(\tau{=}0.15,\; T_a{=}6.4,\; k{=}4)$ with 1000 diffusion steps. A total of $\sim 2 \times 10^5$ steps were used for training (see Appendix.~\ref{app:Cifar} for details). The active version (the generated samples reported have an FID of $7.61$) performs better than its passive counterpart in line with previous findings. We note that this FID score was obtained without any directed finetuning of the active generative diffusion algorithm. The typical tools used to improve FID scores in generative diffusion models, such as using noise scheduling etc, can also be readily applied and ported to the active context. Further, beyond the improvement in performance, as we detail in the next section, our work contributes new understanding for how non-equilibrium correlations have the potential to systematically improve generative diffusion.

\renewcommand*{\thesection}{\Roman{section}}
\section{Active generative diffusion preserves memory traces and class history longer}
\label{sec:ClassStructure}
\renewcommand*{\thesection}{\arabic{section}}

\begin{figure}[ht]
    \centering
    \includegraphics[width=0.95\linewidth]{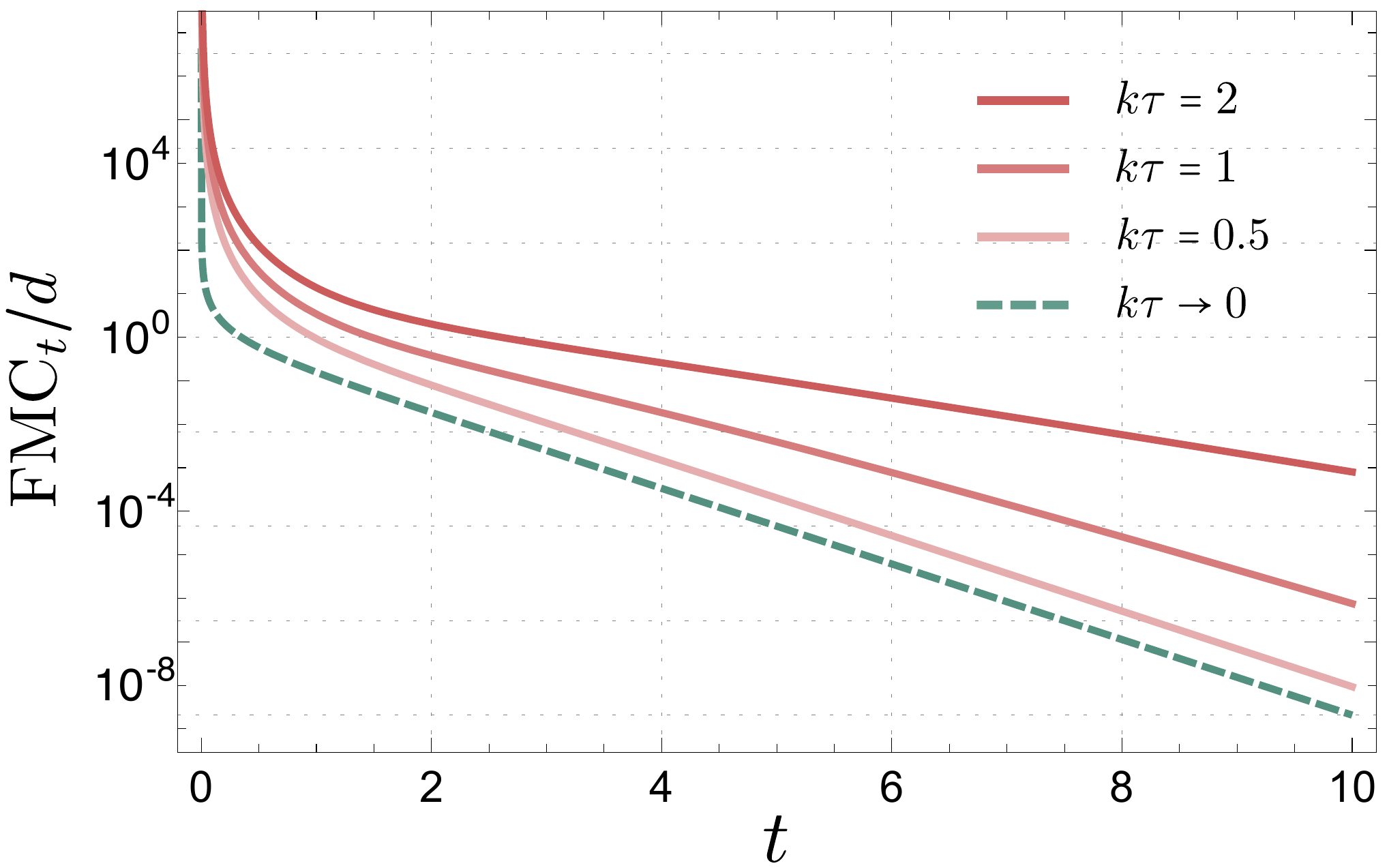}
    \caption{Fisher memory curves $\mathrm{FMC}_t$ for the forward active diffusion process at different activity levels $\tau$, normalized by the dimension of the input data $d$. The passive limit corresponds to the case $\tau \to 0$. As $\tau$ increases, the FMC exhibits a slower decay, indicating that the system retains information from the input for a longer duration. The $y$-axis is shown on a logarithmic scale. Parameters used: $k = 1$, $T_p=0$, and $T_a=1$.\label{fig:FMC} }
\end{figure}

The observations in the previous sections raise a question: 
to what extent and over what time scale is the memory trace of the input data signal preserved in the active forward diffusion process?
To address this question, we compute the Fisher memory curve (FMC), a measure initially introduced for reccurent neural networks in Ref.~\cite{Ganguli-Huh-Sompolinsky08}. 
Defining the state vector $\vec{X}_t = (\mathbf{x}_t^\top,\bm{\eta}_t^\top)^\top \in \mathbb{R}^{2d}$, we consider two different input data: $\vec{X}_0$ and its perturbed version $\vec{X}_0+\delta\vec{X}_0$. 
For the forward process with $t\in(0,t_f]$, the Kullback--Leibler (KL) divergence between two proximate distributions, conditioned on the two different initial inputs, is given by the following quadratic form:
\begin{equation}
    D_{\text{KL}}[P_t(\vec{X}_t|\vec{X}_0+\delta\vec{X}_0) \,\|\, P_t(\vec{X}_t|\vec{X}_0)]
    = \frac{1}{2}\delta\vec{X}_0^\top F_t \delta\vec{X}_0 \ ,
\end{equation}
where $F_t=e^{M^\top t}C_t^{-1}e^{Mt}\in\mathbb{R}^{{2d}\times{2d}}$ defines the spacetime Fisher memory matrix, $M$ and $C_t$ are the Jacobian and the covariance matrix of the forward SDE.
The FMC is defined as the trace of $F_t$ over the spatial indices:
\begin{equation}
    \mathrm{FMC}_t \equiv \mathrm{tr}F_t = \sum_{i=1}^{2d}(F_t)_{ii} \ .
\end{equation}
Notably, the FMC is a quantity independent of the provided input data  ($\vec{X}_0, \vec{X}_0+\delta\vec{X}_0$) and depends solely on the Jacobian $M$ and the noise statistics of the forward SDE under consideration (see e.g. Ref. \cite{Amari16:book}). 
As discussed in Refs.~\cite{Ganguli-Huh-Sompolinsky08, Kerg--Lajoie19}, the FMC thus quantifies how much the system ``remembers" its input over time as an inherent nature of the given dynamical system, independent of the input data statistics.
Fig.~\ref{fig:FMC} compares the FMC for various persistence times $\tau$, where $\tau\to0$ corresponds to the passive (white-noise) limit.
As $\tau$ (and thus the activity) increases, the FMC exhibits a slower decay rate, indicative of a prolonged memory trace of the input signal.


One way to test this hypothesis it to look at how the ``class structure'' of a dataset decays with time. A large number of publicly available datasets -- MNIST, CIFAR-10, ImageNet \textendash to name a few, have labeled data with the labels corresponding to different classes. These classes can be understood as various distinct clusters of points in a high dimensional landscape. We can test our hypothesis by checking the rate at which the clusters collapse into a single clump of pure noise in high dimensions.

\begin{figure}[ht]
    \centering
    \includegraphics[width=0.95\linewidth]{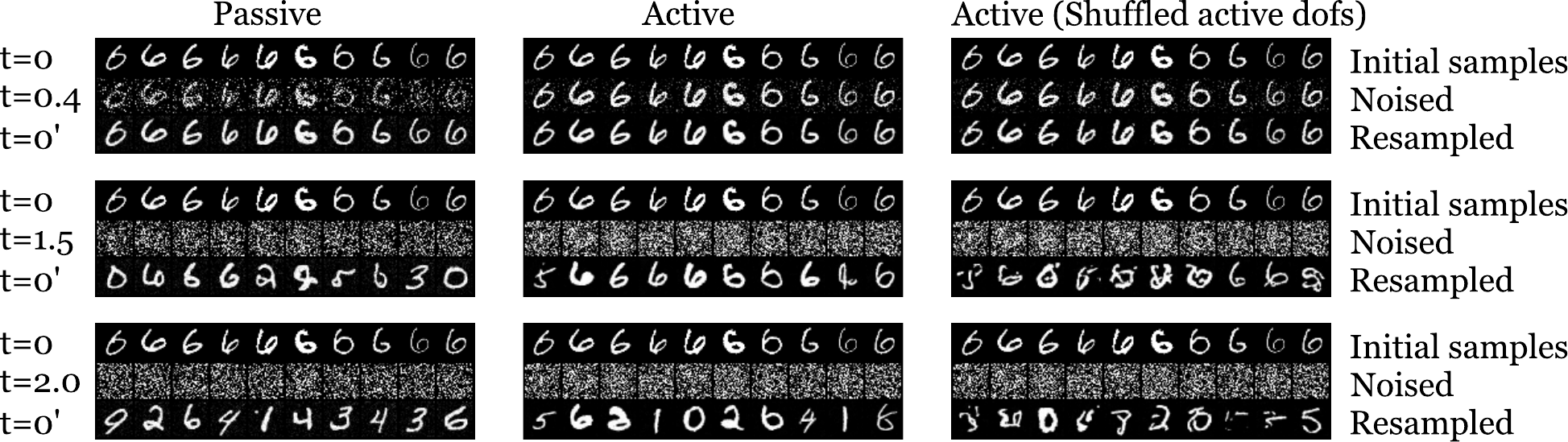}
    \caption{Instance of the numerical experiments carried out. Three different instances of forward diffusion times ($t_f$) are provided. For each of the instances the first row denotes the image samples with which the forward process is initialized with. The middle row denotes the noised image after the initial image is taken through the forward process for time $t$ denoted on the top of the image in the first row. The bottom row denotes the reconstructed image by performing reverse diffusion on the noised image in the middle row. For $t_f=0.4$, the noised images retain the features of the original image (``5'' in this case) and thus the resampling leads to an image from the same class. At long times such features are lost, the resampling process starts from pure noise and leads back to a random class. }
    \label{fig:lostcorrelations}
\end{figure}

\begin{figure}
    \centering
    \includegraphics[width=0.9\linewidth]{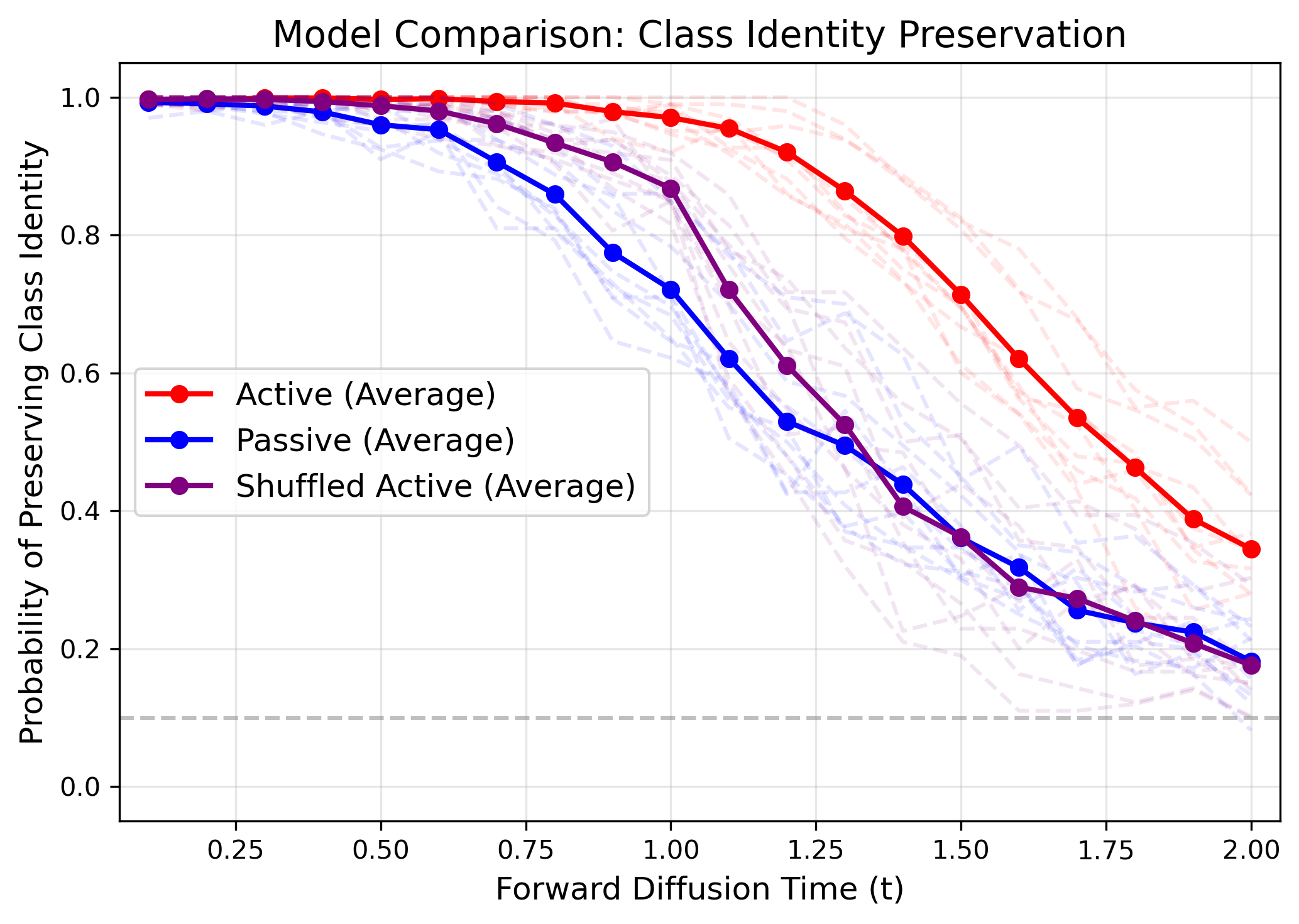}
    \caption{Class recovery curves as a function of forward diffusion times. The blue curve is for the passive process, the red is for the active process. For the purple curve, we scramble the active degrees of freedom at $t_f$ (the start of the resampling process) within the same image. This is to test the hypothesis that the active degrees of freedom $\bm{\eta}$ carry information about the class. The curves show that the passive case loses class information faster than active case and that the active degrees carry a significant amount of information about class structure. The numerical experiments were conducted with 100 randomly chosen images from each of the 10 classes in MNIST i.e. over a total of 1000 images. For the passive process, $T=1.0$ and for the active process, $T_a=1.0, T_p=10^{-3}, \tau=0.5$. For training the neural networks for score function, the total time for forward diffusion was $2.0$.}
    \label{fig:Wyartt_all_models}
\end{figure}

We test for preservation of class structure in the MNIST dataset under the passive and active forward processes. We hypothesize that information about the class is lost at a slower rate in active case than in the passive case. On top of this, the active degrees of freedom (dofs), $\bm{\eta}$, carry a significant part of the information about the class in the active case. To test this, we conduct the following numerical experiment which is inspired from Ref.~\cite{sclocchi2025phase} (see Fig.~\ref{fig:lostcorrelations}). We first learn the score functions for the passive and active processes by training the neural networks as described in Sec.~\ref{sec:ActiveRevTimeDif}. Then we initialize the forward process with a random image from a specific class. We run the forward diffusion for a certain length of time, $t_f$. This destroys the correlations between the pixels of the image and takes it closer to pure noise. We call this the \textit{partially noised image}. Then we reverse the process starting from \textit{this partially noised image} using the score function that was learned earlier. This reverse diffusion process generates a new image which could be from the same or different class that the forward process was initialized with. The classification of the images are performed using a LENet described in the github repository of Ref.~\cite{csinva_mnist}. We plot the statistics of how the fraction of the original recovered class as a function of $t_f$ in Fig.~\ref{fig:Wyartt_all_models}. We observe for the passive case, the class structure is lost rapidly as a function of forward time. For the active case it is much slower, but once we shuffle the active degrees of freedom at the beginning of the resampling process, the class recovery goes down to almost the same level as the passive case. This supports our hypothesis that active dofs carry significant information about the class structure.

\begin{figure*}
    \centering
    \includegraphics[width=0.9\linewidth]{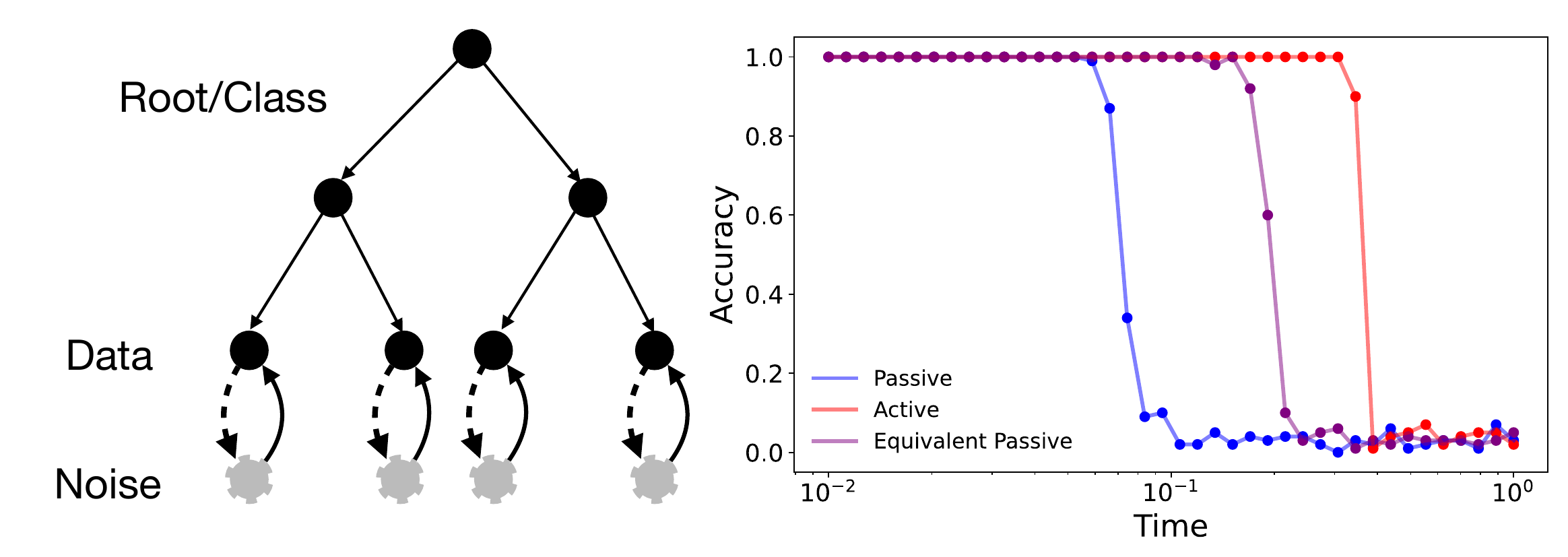}
    \caption{(left) Active dynamics preserves class history for longer in a hierarchical data model. (right) The original class information is preserved for longer with active dynamics. We compare dynamics of a passive model, active model, and a passive model in which the amount of noise added is effectively the same as that in the active model. Importantly the correlation between {\bf x} and $\eta$ can drive recovery even when correlations in {\bf x} are completely destroyed by the noise (as evidenced by the curve labeled ``Equivalent passive"). The common parameters of the hierarchical data structure are, $L=10, m=8, s=2, v=32$. For the various noising processes, parameters used are: Passive - $T=1.0$, Active - $T_a=1.0,T_p=0.001,\tau=2.0$ and for``Equivalent Passive" - $T=1/3$.}
    \label{fig:ActiveHierarchy}
\end{figure*}

To put this idea on a firmer footing, we followed Ref ~\cite{sclocchi2025phase} and analyzed the effect of active noise using a hierarchical data model. Details of the calculation are in Appendix.~\ref{sec:bp_hierarchy}. Specifically, we assume that the observed data at one level of the hierarchy is generated probabilistically from data at the previous level. The data at the lowest level is a proxy for the observed data while data at the highest, or root level, is a proxy for the class the data belongs to. In the context of MNIST images for example, class identity corresponds to the digit identity and finer details of how the digits are drawn are encoded in subsequent levels. We can explore the implication of memory traces due to active noise with these models. We begin by adding noise to the data at the finest level in a mimic forward noising process. We then compute as a function of noising time $t$ the class identity obtained at the root level in a denoising process (similar to the setup in Fig.~\ref{fig:lostcorrelations}. As in Fig.~\ref{fig:lostcorrelations}, we observe that class information is lost sharply in both the passive and active processes (Fig.~\ref{fig:ActiveHierarchy}). Importantly, information is retained longer in the active diffusion process as evidenced by the delayed transition. We note that at points prior to the delayed transition with the active dynamics, reverse dynamics applied on a marginal probability distribution with just the data degrees of freedom fail to recover the class information. In other words, even though the marginal data distribution has lost all information about its priors, this information can be recovered due to the active degrees of freedom. This again reinforces how the active noise degrees of freedom store crucial information about the data degrees of freedom. We note that correlations between ${\bf x}$ and $\eta$, $\langle {\bf x} \cdot{\eta}$ drive so called active pressure or dissipation in active matter systems~\cite{fodor2016far}. Our work shows how these same correlations can help with data generation in generative diffusion. 
These findings are consistent with the empirical observations such as those in Fig.~\ref{fig:multigaussian:NN} where active diffusion is able to generate structure in a smaller number of steps. Indeed, since the active dynamics are able to generate root classes faster its not unreasonable to speculate that they can start to generate a reasonable description of the required data even with a small number of function calls. 

\renewcommand*{\thesection}{\Roman{section}}
\section{Faster speciation with active dynamics}
\label{sec:Discussion}
\renewcommand*{\thesection}{\arabic{section}}

The results of the previous section also suggest an intriguing connection between active dynamics and so called speciation times, i.e. the time at which the first data structures begin to emerge in the reverse diffusion process. Indeed,  from Fig.~\ref{fig:lostcorrelations} we might expect faster speciation times with active processes. Here, following Ref.~\cite{biroli2024dynamical}, we show that with active noise, it takes a shorter amount of time for trajectories in the reverse process to choose their primary class of data.   For instance, in the case of Gaussian mixture model, active diffusion will choose one of the Gaussian basins to fall into sooner than in passive diffusion. The details of the calculation are provided in Sec.~\ref{sec:Speciation}.  Comparing the expression for speciation time with active noise 
$t_s^a = \frac{1}{2} \log(\frac{\rm \max_{\lambda}(C_0)(1+\tau)}{T_a})$, where $\max_{\lambda}(C_0)$ is the largest eigenvalue of the data covariance matrix to the expression for the passive case, $t_s^p = \frac{1}{2} \log(\frac{\rm \max_{\lambda}(C_0)}{T_p})$, we observe that for a fixed target distribution and for the same passive and active temperature ($T_p = T_a$), $t_s^a > t_s^p$ (here the time, $t$, is being measured in the forward diffusion process i.e. $t=0$ corresponds to the data distribution at the start of the forward process). Thus for the reverse diffusion process, the speciation happens faster in the active case when compared to passive case. Faster speciation would imply that more time can be spent sampling the various peaks in the data distribution, which could lead to better fine-scale resolution in the generated configurations. 

\renewcommand*{\thesection}{\Roman{section}}
\section{Conclusion}
\label{sec:Conclusion}
\renewcommand*{\thesection}{\arabic{section}}

In this work, we have explored how driving generative diffusion out of equilibrium with active, correlated noise fundamentally change generative properties. Standard diffusion models rely on passive, memoryless Brownian motion and we argue that this approach can inherently struggle to resolve multi-scale structures in rugged energy landscapes. By coupling the data to auxiliary active degrees of freedom (with some implicit memory) active generative diffusion provides a route to alleviate some of these issues. 

Crucially, we identified physical mechanism driving this enhancement. Our theoretical analysis reveals that active dynamics facilitate an earlier symmetry breaking in the reverse generative process. This allows the system to lock into the correct ``basin of attraction'' (class identity) much earlier than in passive methods, allowing the remainder of the diffusion trajectory to focus on refining fine-grained local fluctuations. This separation of timescales is particularly advantageous for sampling metastable molecular conformations and multi-scale geometries, where passive methods frequently fail to cross high energy barriers. Further, we speculate that the active timescale $\tau$, smooths the temporal variation of the score function, reducing the functional complexity that the neural network must approximate. This suggests that ``physics-informed'' choices in the diffusion process simplify the learning task itself.

Ultimately, our results suggest that the principles of active matter physics can be repurposed to engineer more robust generative models. This opens a new avenue for ``active generative AI'', where the dynamics of learning are tuned not just for algorithmic convergence, but for thermodynamic efficiency in navigating high-dimensional landscapes



\renewcommand*{\thesection}{\Roman{section}}
\section{Acknowledgments}
\label{sec:Acknowledgments}
\renewcommand*{\thesection}{\arabic{section}}
SV, CF and AL acknowledge support by the National Institute of General Medical Sciences of the NIH under Award No. R35GM147400. AKB acknowledges support from a fellowship from the Department of Chemistry at the University of Chicago.  We acknowledge support from the National Science Foundation through the Physics Frontier Center for Living Systems (PHY-2317138). AN gratefully acknowledges support from the Eric and Wendy Schmidt AI in Science Postdoctoral Fellowship, a Schmidt Sciences, LLC program. 

\bibliographystyle{unsrt}
\bibliography{references}

\clearpage

\begin{widetext}

\setcounter{figure}{0}
\setcounter{table}{0}
\setcounter{equation}{0}
\setcounter{page}{1}
\setcounter{section}{0}

\renewcommand{\theequation}{A\arabic{equation}} 
\renewcommand{\thepage}{A\arabic{page}} 
\renewcommand{\thesection}{A\arabic{section}} 
\renewcommand{\thesubsection}{\arabic{subsection}}
\renewcommand{\thetable}{A\arabic{table}}  
\renewcommand{\thefigure}{A\arabic{figure}}

\section{Derivation for Active reverse diffusion}
\label{appsec:Derivation}
The forward active diffusion process for $\mathbf{x}(t),\bm{\eta}(t)\in\mathbb{R}^d$ is defined, for $t\in[0,t_f]$, by the system of stochastic differential equations (SDEs)
\begin{align}
  \dot{\textbf{x}} =& -k\textbf{x} + \bm{\eta}(t) + \bm{\xi}_1(t) \ ,  \label{eq:fd-x}\\
  \dot{\bm{\eta}} =& -\frac{\bm{\eta}}{\tau} + \bm{\xi}_2(t) \ ,
\end{align} 
where $\bm{\xi}_1,\bm{\xi}_2$ are independent Gaussian white noise with, for all $i,j\in\{1,\dots,d\}$,
\begin{align}
\langle \xi_{1,i}(t) \rangle = 0 \ , & \quad \langle \xi_{2,i}(t) \rangle = 0 \ ,  \\
  \langle \xi_{1,i}(t) \xi_{1,j}(t') \rangle &= 2T_p\delta_{ij} \delta(t-t') \ , \\
  \langle \xi_{2,i}(t) \xi_{2,j}(t') \rangle &= \frac{2T_a}{\tau^2} \delta_{ij} \delta(t-t') \   \label{eq:fd-xi2}.
\end{align}
Here the angle bracket $\langle\cdots\rangle$ denotes the ensemble average over the noise realizations $(\bm{\xi}_1,\bm{\xi}_2)$.
Unless otherwise stated, we assume $k\tau\neq1$ throughout the paper.

The initial condition for the forward process is the joint distribution between the data degrees of freedom ($\textbf{x}$) and the corresponding active degrees of freedom ($\bm{\eta}$), and it is constructed as 
\begin{align}
   P_0(\textbf{x}_0, \bm{\eta}_0) = P_0(\textbf{x}_0) P_0(\bm{\eta}_0) \label{eq:ActiveIC}
\end{align}
where we assume 
\begin{equation}
    P_0(\bm{\eta}_0) = \mathcal{N}(\bm{\eta}_0; \mathbf{0}, \tau^{-1}T_aI_d) \propto \exp(-\frac{\tau}{2T_a}\lVert\bm{\eta}_0\rVert^2)
    \label{eq:P_eta0}
\end{equation}
and $P_0(\textbf{x}_0)$ is the distribution from which the data is drawn. Here $\lVert\cdot\rVert$ denotes the standard Euclidean norm. With these initial conditions, the conditional distribution $P(\textbf{x},\bm{\eta}|\mathbf{x}_0,\bm{\eta}_0;t)$ is given by (see, Sec.~\ref{appsec:PerturbData})
\begin{align}
  P(\textbf{x},\bm{\eta}| \textbf{x}_0, \bm{\eta}_0;t)
  = \mathcal{N}(\vec{X}_t;\,\vec{\mu}_t, C_t)
  \propto \exp[- \frac{(\vec{X}_t-\vec{\mu}_t)^\top C_t^{-1}(\vec{X}_t-\vec{\mu}_t)}{2}] \label{eq:ActiveConditionalP} 
\end{align}
where we have defined the state vector $\vec{X}_t=\begin{pmatrix}\mathbf{x}_t\\\bm{\eta}_t\end{pmatrix}\in\mathbb{R}^{2d}$ and
\begin{align}
  \vec{\mu}_t &= \begin{pmatrix}  e^{-kt}\textbf{x}_0 + \frac{e^{-t/\tau} - e^{-kt}}{k-\frac{1}{\tau}}\bm{\eta}_0 \\  e^{-t/\tau}\bm{\eta}_0 \end{pmatrix}
  \label{eq:mu_t}
  \\
  C_t &= \begin{pmatrix} m_{11} \ m_{12} \\ m_{12} \ m_{22} \end{pmatrix} \otimes I_d \\
  m_{11} &= \frac{T_p}{k}(1-a^2) + \frac{T_a}{\tau^2}\left( \frac{\tau}{kc} + \frac{1}{d^2} \left( \frac{4ab}{c} - b^2\tau -\frac{a^2}{k} \right) \right) \label{eq:m11}\\
  m_{12} &= \frac{T_a}{\tau cd} \left( k(1-b^2) -\frac{1}{\tau} \left( 1+b^2-2ab \right) \right) \label{eq:m12}\\
  m_{22} &= \frac{T_a}{\tau} (1-b^2) \label{eq:m22}\\
  a &= e^{-kt}, \quad b=e^{-t/\tau}, \quad c = k+\frac{1}{\tau}, \quad d = k-\frac{1}{\tau} \ ,
  \label{eq:abcd}
\end{align}
where $\cdot^\top$ denotes the transpose operation, $\otimes$ the Kronecker product, and $I_d$ the identity in $\mathbb{R}^{d\times d}$.
Marginalizing the conditional probability on the initial distribution $P_0(\mathbf{x}_0, \bm{\eta}_0)=P_{0}(\bm{x}_0)P_{0}(\bm{\eta}_0)$ gives the unconditioned distribution at time $t$:
\begin{align}\label{eq:unconditioned_P}
    P(\mathbf{x}, \bm{\eta}; t) = \int P(\mathbf{x}, \bm{\eta} | \mathbf{x}_0, \bm{\eta}_0; t) P_0(\mathbf{x}_0, \bm{\eta}_0) \,d\mathbf{x}_0d\bm{\eta}_0 \ .
\end{align}

One admissible reverse diffusion SDE for this process is given by (\cite{anderson1982reverse, Song_2021_SGM_SDE})
\begin{align}
  -\dot{\textbf{x}} =& -k\textbf{x} + \bm{\eta} + 2T_p \mathscr{F}_\textbf{x}(\textbf{x},\bm{\eta};t) + \bm{\xi}_1(t) \\
  -\dot{\bm{\eta}} =& -\frac{\bm{\eta}}{\tau} + \frac{2T_a}{\tau^2} \mathscr{F}_{\bm{\eta}}(\textbf{x},\bm{\eta};t) + \bm{\xi}_2(t)
\end{align}
where $\mathscr{F}_{\textbf{x}}(\textbf{x},\bm{\eta};t)\equiv  \nabla_{\textbf{x}} \log P(\textbf{x},\bm{\eta};t)$ and $\mathscr{F}_{\bm{\eta}}(\textbf{x},\bm{\eta};t)\equiv  \nabla_{\bm{\eta}} \log P(\textbf{x},\bm{\eta};t)$ are the score functions for this process.

Using Eq.~\eqref{eq:unconditioned_P}, we have
\begin{align}
    \begin{pmatrix}
    \mathscr{F}_{\bm{x}}(\mathbf{x},\bm{\eta};t)\\
    \mathscr{F}_{\bm{\eta}}(\mathbf{x},\bm{\eta};t)
    \end{pmatrix}
    &= \frac{1}{P(\mathbf{x}, \bm{\eta}; t)} \int\big[\!-C_t^{-1}\vec{X}\big]
    P(\mathbf{x}, \bm{\eta} | \mathbf{x}_0, \bm{\eta}_0; t) P_0(\mathbf{x}_0, \bm{\eta}_0) \,d\mathbf{x}_0d\bm{\eta}_0
    \nonumber
    \\
    &= \int \big[\!-C_t^{-1}\vec{X}\big]P(\mathbf{x}_0, \bm{\eta}_0 | \mathbf{x}, \bm{\eta}; t) \,d\mathbf{x}_0d\bm{\eta}_0 \ ,
\end{align}
where we have defined the conditional posterior distribution $P(\mathbf{x}_0, \bm{\eta}_0 | \mathbf{x}, \bm{\eta}; t)=P(\mathbf{x}, \bm{\eta} | \mathbf{x}_0, \bm{\eta}_0; t) P_0(\mathbf{x}_0, \bm{\eta}_0)/P(\mathbf{x}, \bm{\eta}; t)$ based on the Bayes' rule.
We then obtain the expressions for the loss functions corresponding to the scores:
\begin{align}
    \mathscr{F}_\textbf{x}(\textbf{x},\bm{\eta};t) &= \frac{-m_{22}(\textbf{x} - a \braket{\textbf{x}_0}_t - \frac{a-b}{d}\braket{\bm{\eta}_0}_t) + m_{12}(\bm{\eta} - b \braket{\bm{\eta}_0}_t)}{\Delta_t} \\
    \mathscr{F}_{\bm{\eta}}(\textbf{x},\bm{\eta};t) &= \frac{-m_{11}(\bm{\eta} - b\braket{\bm{\eta}_0}_t) + m_{12}(\textbf{x} - a \braket{\textbf{x}_0}_t - \frac{a-b}{d}\braket{\bm{\eta}_0}_t)}{\Delta_t} \label{eq:ScoreMatching}
\end{align}
where $\Delta_t\equiv\det\begin{pmatrix}m_{11}&m_{12}\\m_{12}&m_{22}\end{pmatrix}=m_{11}m_{22}-m_{12}^2$ and $\braket{\cdots}_t$ denotes the expectation over the backward conditional distribution $\mathbb{E}_{(\mathbf{x}_0,\bm{\eta}_0)\sim P(\cdot|\textbf{x},\bm{\eta};t)}[\cdots]=\int d\mathbf{x}_0d\bm{\eta}_0\cdots P(\mathbf{x}_0,\bm{\eta}_0|\mathbf{x},\bm{\eta};t)$.

Setting $T_p=0$ in the active process allows us to ignore $\mathscr{F}_{\textbf{x}}$ and learn only $\mathscr{F}_{\bm{\eta}}$. See Ref.~\cite{dockhorn2021score} for a discussion on why $\mathscr{F}_{\textbf{x}}$ is more difficult to learn.

\section{Numerical Implementation}
\label{appsec:NumericalImplementation}

\subsection{Extension of the CLD~\cite{dockhorn2021score} Code}
\label{appsec:NumericalImplementation:CLD_extension}

Here we discuss the numerical implementation of reverse diffusion process using a score objective and the architectures of the neural networks used. We adapt the implementation from Ref.~\cite{dockhorn2021score}. First, the initial data, $(\textbf{x}_0, \bm{\eta}_0) \sim P(\textbf{x}_0, \bm{\eta}_0)$, are perturbed in the forward process over a time interval, $t \sim \mathscr{U}[0,t_f]$, where $\mathscr{U}$ is the uniform distribution and $t_f$ is the length of time for which the forward process is run. The perturbation kernel for active diffusion is given in Sec.~\ref{appsec:PerturbData}. For $t_f \to \infty$, the data distribution reduces to a multidimensional Gaussian distribution centered at the origin. Using these generated samples at different times, we train the neural network to minimize the score objective, a hybrid score matching objective in our case~\cite{vahdat2021score}. To ``inform'' the neural network about the structure of the score, we use a mixed score parameterization~\cite{vahdat2021score, dockhorn2021score} (Sec.~\ref{appsec:hybrid_score_matching}).

The primary difference between CLD and active diffusion is the stochastic differential equations (SDEs) defining each process. Thus, implementing active diffusion was achieved by sub-classing the existing CLD class and overriding key methods such as the SDEs of the diffusion process and the perturbation kernel functions. 

\subsection{Perturbation Kernel for Active Diffusion}
\label{appsec:PerturbData}
Denoting the state vector by $\vec{X}_t = (\mathbf{x}_t^\top\!, \bm{\eta}_t^\top)^\top \in \mathbb{R}^{2d}$, the forward active diffusion process \eqref{eq:fd-x}--\eqref{eq:fd-xi2} can be recast in the following form:
\begin{align}\label{eq:recasted_forward_diffusion}
    d\vec{X}_t &= M\vec{X}_tdt + Gd\vec{w}_t \\
    M &= \begin{pmatrix}
        -k & 1 \\
        0 & -1/\tau
    \end{pmatrix} \otimes I_d \ , \ 
    G = \begin{pmatrix}
        \sqrt{2T_p} & 0 \\
        0 & \sqrt{2T_a}/\tau
    \end{pmatrix} \otimes I_d \ ,
\end{align}
where $\vec{w}_t$ is a standard Wiener process in $\mathbb{R}^{2d}$. The solution to this equation can be formally written as
\begin{align}
    \vec{X}_t &= e^{Mt}\vec{X}_0 + \int_0^t e^{M(t-s)} G \,d\vec{w}_s \ .
\end{align}
Denoting the expectations with respect to the Wiener noise by $\langle\cdots\rangle$, we have $\langle\vec{X}_t\rangle=e^{Mt}\vec{X}_0$ and
\begin{equation}
  \langle{\vec{X}_t\vec{X}_t^\top}\rangle = e^{Mt}{\vec{X}_0\vec{X}_0^\top} e^{M^\top t}
  \;+\;\int_0^t {e}^{Ms}\, G G^\top{e}^{M^\top s}\,{d}s \ .
\end{equation}
Since in the SDE \eqref{eq:recasted_forward_diffusion} the drift is linear and the noise is additive Gaussian, if the data is normally distributed at $t=0$, it is normally distributed throughout the entire forward process ($t\in[0,t_f]$). 
Expanding the expressions above, the mean ($\vec{\mu}_t=\langle{\vec{X}_t}\rangle$) and covariance ($C_t=\langle{(\vec{X}_t-\vec{\mu}_t)(\vec{X}_t-\vec{\mu}_t)^\top}\rangle=\langle{\vec{X}_t\vec{X}_t^\top}\rangle-\vec{\mu}_t\vec{\mu}_t^\top$) of the data at various time instants in the forward process are then given by Eqs.~\eqref{eq:mu_t}--\eqref{eq:abcd}.

We now set $T_p=0$.
Marginalizing the conditional distribution $P(\mathbf{x},\bm{\eta}|\mathbf{x}_0,\bm{\eta}_0;t)$ \eqref{eq:ActiveConditionalP} over the initial distribution of active degrees of freedom $P_0(\bm{\eta}_0)$ \eqref{eq:P_eta0}, we obtain the partially-conditional distribution
\begin{equation}
    P(\vec{X}_t|\mathbf{x}_0;t) 
    = \mathcal{N}(\vec{X}_t;\vec{\bar{\mu}}_t,\bar{C}_t) 
    \propto \exp\bigg[ -\frac{(\vec{X}_t-  \vec{\bar{\mu}}_t)^\top \bar{C}_t^{-1} (\vec{X}_t - \vec{\bar{\mu}}_t)}{2} \bigg]
\end{equation}
where the mean vector and the covariance matrix are given respectively by
\begin{align}
    \vec{\bar{\mu}}_t &= \begin{pmatrix} e^{-kt}\textbf{x}_0  \\ \mathbf{0} \end{pmatrix} \\
    \bar{C}_t &= \begin{pmatrix} \bar{m}_{11} \ \bar{m}_{12} \\ \bar{m}_{12} \ \bar{m}_{22} \end{pmatrix} \otimes I_d \ 
\end{align}
with
\begin{align}
    \bar{m}_{11} &= \frac{T_a}{\tau}\bigg( \frac{1-a^2}{kc} - \frac{2}{cd}(ab-a^2) \bigg) = \frac{T_a}{\tau}\bigg(\frac{1}{kc}+\frac{a^2}{kd}-\frac{2}{cd}ab\bigg) \\
  \bar{m}_{12} &= \frac{T_a}{\tau c} \left( 1 - ab \right) \\
  \bar{m}_{22} &= \frac{T_a}{\tau}  \\
  a &= e^{-kt} , \ b=e^{-t/\tau} , \ c = k+\frac{1}{\tau}, \ d = k-\frac{1}{\tau} \ .
\end{align}
Here $\bar{\cdot}$ represents the partially-marginalized quantities. 

The reason for this partial marginalization will become clear in the next section where we use hybrid score matching (HSM) for the active process~\cite{vahdat2021score, dockhorn2021score}.

\subsection{Hybrid Score Matching and Mixed Score Parametrization}
\label{appsec:hybrid_score_matching}

For the 2D Swiss Roll, multiple Gaussians and 2D alanine dipeptide datasets, we use the hybrid score matching objective for training the neural network. For the 25D alanine dipeptide and 2D Ising model datasets, we use the score-mixing objective for training (see Sec.~\ref{appsec:NumericalImplementation}).

The score matching objective that the neural network needs to optimize is given in Eq.~\ref{eq:ScoreMatching}, 
\begin{align}
    \mathscr{L}(w) =& \mathbb{E}_{t\sim \mathscr{U}[0,t_f], \vec{X}(t) \sim P(\vec{X}(t))}\left[ ||S_\textbf{w}^{(\bm{\eta})}(\vec{X}) - \mathscr{F}_{\bm{\eta}}(\vec{X}(t))||^2 \right] \\
    =& \mathbb{E}_{t\sim \mathscr{U}[0,t_f], \vec{X}(t) \sim P(\vec{X}(t))}\left[ ||S_\textbf{w}^{(\bm{\eta})}(\vec{X}) - \nabla_{\bm{\eta}}P(\vec{X}(t))||^2 \right].
\end{align}

Ref.~\cite{vahdat2021score} showed that for a forward process conditioned on additional degrees of freedom apart from the data, this loss can be replaced with an equivalent function which makes use of the distribution of the additional variables (active degrees of freedom in our case). The new score function is the HSM function, given by
\begin{align}
    \mathscr{L}_{\rm HSM}(w) =& \mathbb{E}_{t \sim \mathscr{U}[0, t_f], \textbf{x}_0 \sim P(\textbf{x}_0), \vec{X} \sim P(\vec{X}(t)|\textbf{x}_0)} \left[ || S_\textbf{w}^{(\bm{\eta})}(\vec{X}) - \nabla_{\bm{\eta}} P(\vec{X}(t)|\textbf{x}_0) ||^2 \right] \label{eq:HSM}
\end{align}

We parameterize this score objective (Eq.~\eqref{eq:HSM}), further following Ref.~\cite{dockhorn2021score}, as
\begin{align}
    S_w^{(\bm{\eta})}(\vec{X}) = -\frac{\bm{\eta}}{m_{22}} + S_\textbf{w}^{\rm new (\bm{\eta})}(\vec{X})
\end{align}
where $m_{22}$ is the element of the covariance matrix representing the variance in the active degrees of freedom, $\bm{\eta}$. $S_\textbf{w}^{\rm new (\bm{\eta})}(\vec{X})$ is the new parameterization of score objective that the neural network has to learn. This parameterization ensures that the neural network tries to learn only the reverse process for the active degrees since the reverse process for the data degrees is a deterministic process entirely determined by the active degrees.

\subsection{Description of Datasets}

The 2D alanine dipeptide toy model was implemented by loading in a file with training data and randomly drawing from the imported dataset. For diffusion on the Ising model, each training sample, a 32-by-32 pixel image with 1 channel, was generated by performing 1,000,000~MCMC sampling steps from a random initial configuration. The training data was generated to have discrete values ($-1$ or $+1$), the perturbation kernel was allowed to vary lattice sites in a continuous manner, and the final samples were discretized back to values of $(-1,+1)$ using a cutoff of $0$. 

\subsection{Neural Network Architectures and Model Training Details}
\label{subsec:NNarch}

In the toy model examples (Gaussian mixtures, Swiss rolls, and 2D alanine dipeptide), the score was learned by a multi-layer perceptron with 4 hidden layers of 128 nodes each. The 25D alanine dipeptide datasets and Ising lattices were treated as images with one channel, and the score was learned using a neural network with the NCSN++ architecture~\cite{Song_2021_SGM_SDE}.

For the toy models, the batch size for each iteration was $512$ training samples; for Ising lattices, the batch size was $1$ sample per iteration. After training, all models were used to synthesize 10,000 samples. 

\subsection{Sampling}
\label{appsubsec:sampling}

All toy models (excepting the 2D alanine dipeptide model) used the Euler-Maruyama sampling scheme. The 2D and 25D alanine dipeptide and Ising models used an ODE sampler with adaptive step size. 

In the 2D toy model examples, we disable denoising at the last step of the sampling scheme. Previous diffusion studies observed that a denoising step (in which only the drift term of the reverse SDE was applied, and not the diffusion term) improved the FID scores of generated image samples by removing noise that is otherwise undetectable by the human eye~\cite{Jolicoeur-Martineau_2021}. This positive effect of the last denoising step is most evident in passive diffusion and does not affect the quality of the samples generated by CLD (see Ref.~\cite{dockhorn2021score}) or, by extension, active diffusion. We disabled denoising in toy models to be able to directly compare the performance of analytic and numeric score functions (Fig.~\ref{fig:multigaussian}), but retain denoising in the 2D alanine dipeptide and Ising model diffusion since these datasets are image-like in nature. 

Although we make direct comparisons of the performance of sample generation of active diffusion and CLD using the EM and ODE samplers, we note that CLD was found to perform best using a custom sampling scheme created by the authors of the method~\cite{dockhorn2021score}. The comparison of performance of active diffusion and CLD using this sampler is intended in future iterations of this work.

\section{Gaussian Mixtures and Swiss Rolls \label{appsec:toy_models}}

\subsection{Reverse Diffusion with Analytic Score on Mixtures of Gaussians}
\label{appsec:AnalyticReverseDiffusion}


The passive forward diffusion process leads to the evolution of the probability density in the following way,
\begin{align}
    P_t(\mathbf{x}) & \propto \sum_{\alpha} \frac{p_{\alpha}}{\Pi_{i}\sqrt{h_i^{\alpha}}} \int Dx_{0,i} \exp \left(-\sum_{i} \left( \frac{(x_{0,i} - \mu_i^{\alpha})^2}{2h_i^{\alpha}} + \frac{(x_i-ax_{0,i})^2}{2\Delta} \right) \right) \\
    & = \sqrt{\Delta}\sum_{\alpha} \frac{p_{\alpha}}{\Pi_{i}\sqrt{\Delta + h_i^{\alpha}a^2}} \exp \left( -\sum_{i} \frac{(x_i-a\mu_i^{\alpha})^2}{2(\Delta + a^2 h_i^{\alpha})} \right)
\end{align}
where $a = e^{-kt}$, the index $i$ runs over the different peaks, and the index $\alpha$ runs over the dimensions of the dataset, and $\Delta$ is given as, $\Delta = \frac{T}{k}(1-e^{-2kt})$. Thus the score function for the reverse process is given by
\begin{align}
    \frac{\partial \log(P_t(\mathbf{x}))}{\partial x_{i}} =& -\frac{1}{P_t(\mathbf{x})}\sum_{\alpha} \frac{p_{\alpha}}{\Pi_{j}\sqrt{\Delta + h_j^{\alpha}a^2}} \frac{(x_{i} - a\mu_i^{\alpha})}{\Delta + a^2 h_i^{\alpha}}\exp\left( -\sum_{j} \frac{(x_{j}-a\mu_j^{\alpha})^2}{2(\Delta + a^2 h_j^{\alpha})} \right)
\end{align}

Following the same procedure for the active process yields
\begin{align}
  P_0(\mathbf{x}_{0}) & \propto \sum_{\alpha} p_{\alpha} \prod_{i} \left[ \frac{1}{\sqrt{h_i^{\alpha}}} \exp \left(-\frac{(x_{0,i}-\mu_i^{\alpha})^2}{2h_i^{\alpha}} \right) \right] \ , \ P_0(\bm{\eta}_0) \propto  \exp \left(-\frac{\bm{\eta}_0^2}{2g} \right) \\
  P(\bm{x},\bm{\eta}| \bm{x}_0, \bm{\eta}_0;t) & \propto \exp \left(- \frac{\vec{X}^T C^{-1} \vec{X}}{2} \right) \ , \ \vec{X} = \begin{pmatrix} \bm{x} - a\bm{x}_0 - b\bm{\eta}_0 \\ \bm{\eta} - c\bm{\eta}_0 \end{pmatrix} \ , \ C = \begin{pmatrix} m_{11} \ m_{12} \\ m_{12} \ m_{22} \end{pmatrix}
\end{align}

\begin{align}
  P_t(\bm{x},\bm{\eta}) & \propto \sqrt{\Delta} g^{dims}\sum_{\alpha} p_{\alpha}  \prod_{i} \left( \frac{h_i^{\alpha}}{\sqrt{\Delta_{\rm eff, i}^{\alpha}}} \right) \prod_{i} \left( \exp \left(-\frac{k_1(x_{i}-a\mu_i^{\alpha})^2 - 2k_2(x_{i}-a\mu_i^{\alpha})\eta_{i}+ k_{3,i}^{\alpha}\eta_{i}^{2}}{2\Delta_{\rm eff, i}^{\alpha}} \right) \right) \\
  k_1 &= c^2 g + m_{22} \ , \ k_2 = bcg + m_{12} \ , \ k_{3,i}^{\alpha} = b^2g+a^2h_i^{\alpha}+m_{11} \ , \\ 
  \Delta_{\rm eff, i} &= k_1k_{3,i}^{\alpha} - k_2^2 \ , \ \Delta = m_{11}m_{22} - m_{12}^2
\end{align}

\begin{align}
  \frac{\partial \ln(P(\bm{x},\bm{\eta}))}{\partial x_{i}} &= -\frac{1}{P_t(\bm{x},\bm{\eta})} \sum_{\alpha} \prod_{j} \left( \frac{h_j^{\alpha}}{\sqrt{\Delta_{\rm eff, j}^{\alpha}}} \right) p_{\alpha} \frac{k_1(x_{i} - a \mu_i^{\alpha})- k_2 \eta_{i}}{\Delta_{\rm eff, i}^{\alpha}} z_{\alpha} \\
  \frac{\partial \ln(P(\bm{x},\bm{\eta}))}{\partial \eta_{i}} &= -\frac{1}{P_t(\bm{x},\bm{\eta})} \sum_{\alpha} \prod_{j} \left( \frac{h_j^{\gamma}}{\sqrt{\Delta_{\rm eff, j}^{\gamma}}} \right) p_{\alpha} \frac{k_{3,i}^{\alpha}\eta_{i} - k_2 x_{i}}{\Delta_{\rm eff, i}^{\alpha}} z_{\alpha} \\
  z_{\alpha} &= \exp \left(-\sum_{r} \left( \frac{k_1(x_{r}-a\mu_r^{\alpha})^2 - 2k_2(x_{r}-a\mu_r^{\alpha})\eta_{r}+ k_{3,r}^{\alpha}\eta_{r}^{2}}{2\Delta_{\rm eff, r}^{\alpha}} \right) \right). 
\end{align}

Fig.~\ref{si:fig:analytic_score} illustrates the effect of reverse diffusion step size on sample synthesis quality when the analytic score function is used. Two datasets are presented, both consisting of 9 Gaussian peaks with standard deviation $\sigma=0.04$ and differing in the spacing between the peaks. The positions of the means of the Gaussian peaks $(\mu_x^i, \mu_y^i)$ is given by,
\begin{align}
    \left[ (\mu_x, \mu_y) \right] = \left[ (0,0), (r, 0), \left(\frac{r}{\sqrt{2}}, \frac{r}{\sqrt{2}}\right), (0, r), \left(-\frac{r}{\sqrt{2}}, \frac{r}{\sqrt{2}} \right), (-r,0), \left(-\frac{r}{\sqrt{2}}, -\frac{r}{\sqrt{2}}\right), (0, -r), \left(\frac{r}{\sqrt{2}}, -\frac{r}{\sqrt{2}}\right) \right]
\end{align}
where the value of $r$ sets the distance between individual Gaussian peaks. In Figs.~\ref{si:fig:analytic_score:diamond} and \ref{si:fig:analytic_score:diamond_close}, $r=1/\sqrt{2}$ and $r=\sqrt{2}/5$, respectively. For the larger spacing between peaks ($r=1/\sqrt{2})$, passive performance is comparable to active for $dt=0.002$. At smaller peak separation, however, passive diffusion does not resolve the peaks as well as active diffusion at the same value of the smallest examined $dt$.

\begin{figure*}[htb]
    \begin{subfigure}[t]{0.75\textwidth}
    \centering
    \caption{(a) Diamond of Gaussians}
    \label{si:fig:analytic_score:diamond}
    \includegraphics[width=0.95\linewidth]{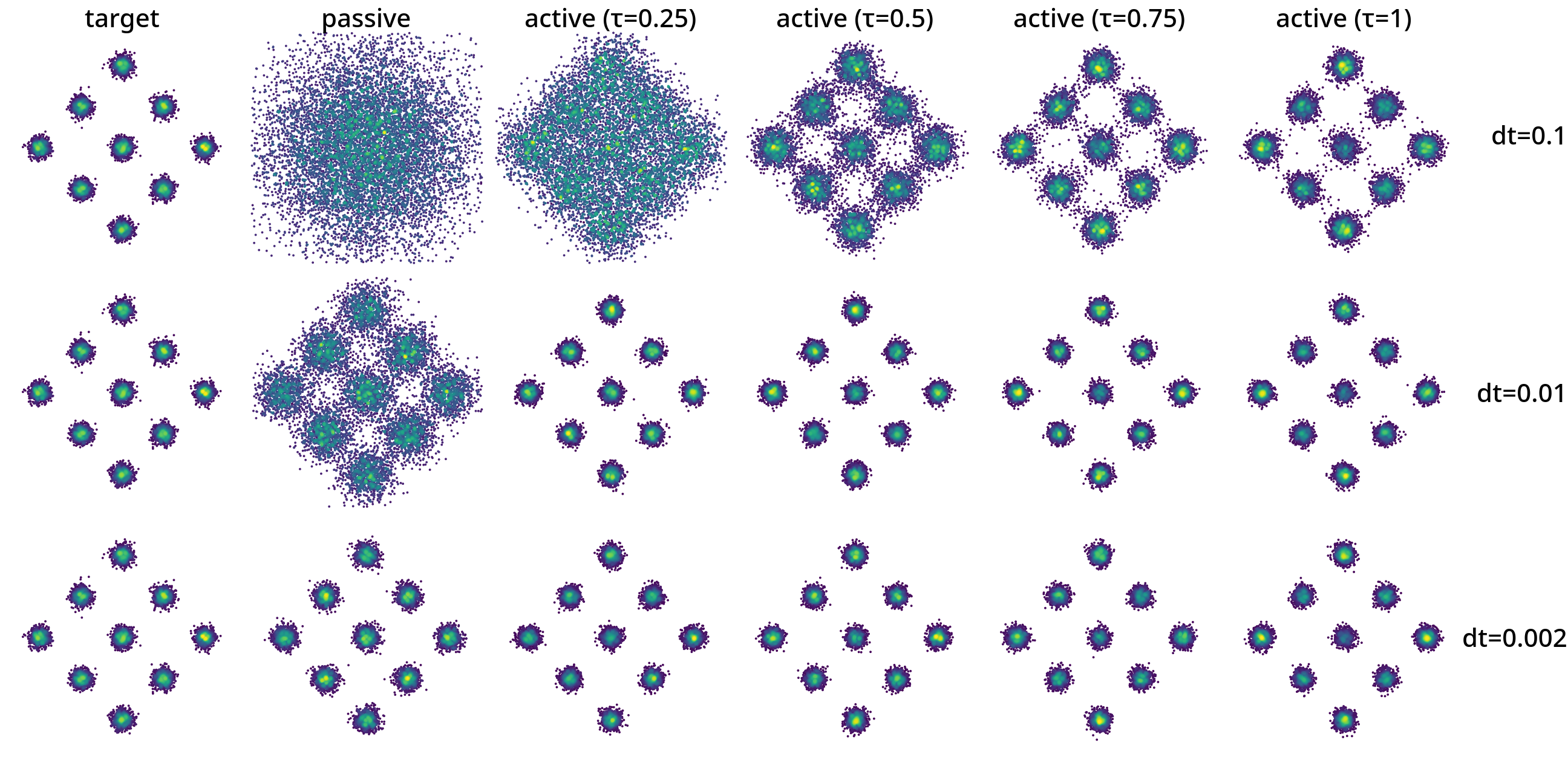}
    \end{subfigure}%
~
    \vskip\baselineskip
    \begin{subfigure}[t]{0.75\textwidth}
    \centering
    \caption{(b) Diamond of Gaussians, closer spacing between peaks}
    \label{si:fig:analytic_score:diamond_close}
    \includegraphics[width=0.95\linewidth]{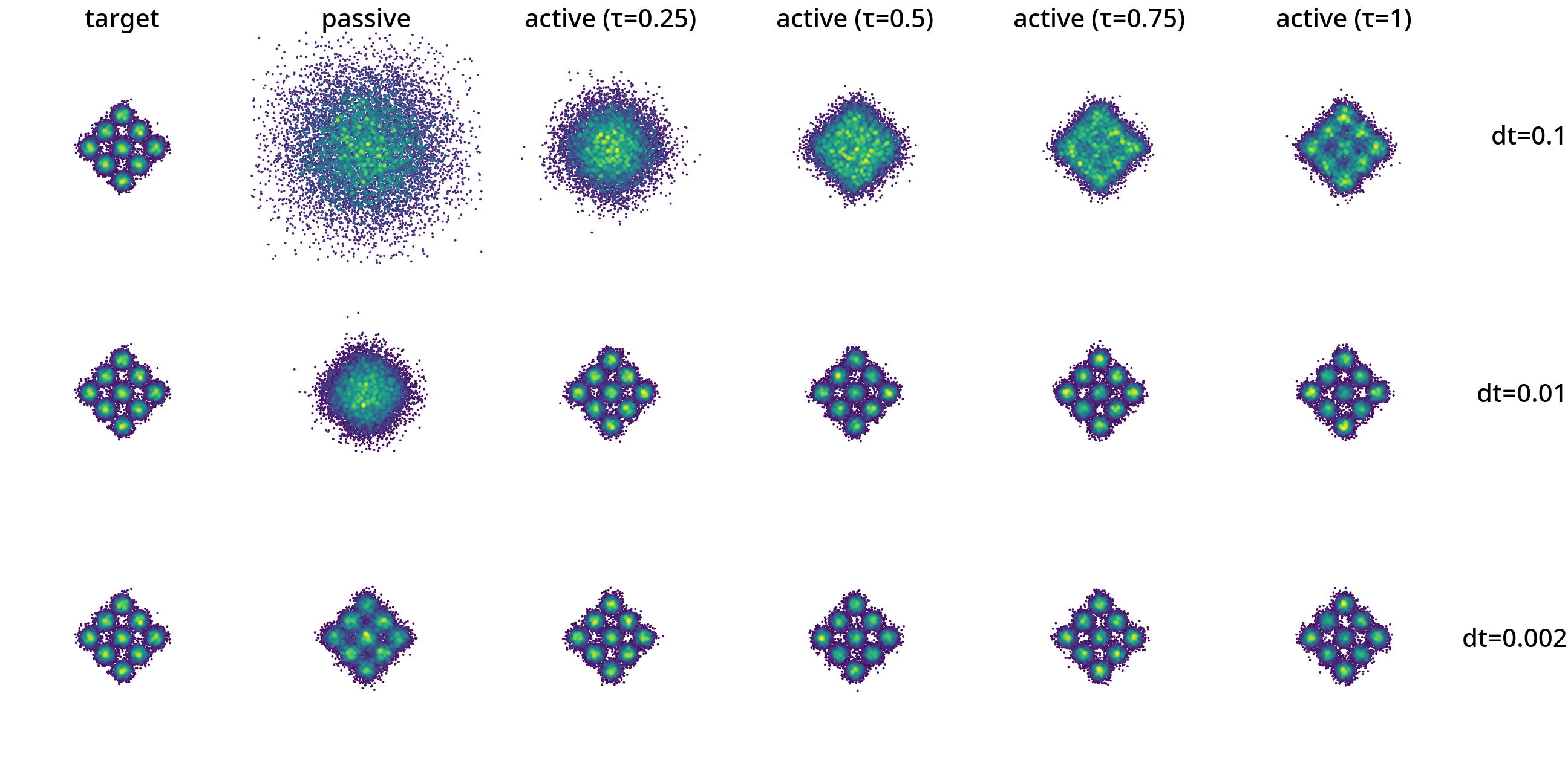}
    \end{subfigure}%
~
    \caption{Analytic score function for two sets of 9 Gaussian peaks with standard deviation $\sigma=0.04$ but with different spacings between peaks. Gaussian peaks with spacing (a) $r=1/\sqrt{2}$ and (b) $r=\sqrt{2}/5$. Increasing the correlation time $\tau$ improves the performance of active diffusion, which is most evident at the largest $dt$ examined here. At large $dt$, active diffusion outperforms passive diffusion. (a) At $dt=0.002$, passive and active diffusion resolve the individual peaks to a similar degree. (b) Decreasing the spacing between peaks also decreases the resolution of peaks for passive diffusion for $dt=0.002$, while active diffusion is not as affected by the change in distance between peaks.}
    \label{si:fig:analytic_score}
\end{figure*}

\subsection{Numerical Diffusion on Gaussian and Swiss Roll Toy Models}

Here we examine the performance of passive, CLD, and active diffusion for a variety of 2D toy model distributions. For each distribution, we change the time step $dt$ of the reverse diffusion process and compare the diffusion generated samples to the target distribution. As mentioned before, denoising is turned off in the last step for these toy models (see Sec.~\ref{appsubsec:sampling}).

Fig.~\ref{si:fig:multigaussian} examines the same distributions that were tested with the analytic score model (Sec.~\ref{sec:analytic_score} and Sec.~\ref{appsec:AnalyticReverseDiffusion}). Passive diffusion  begins to resolve the positions of the larger features at $dt=0.1$. CLD and active diffusion  perform better than passive diffusion for $dt=0.01$ and $dt=0.002$.

\begin{figure*}[htb]
     \begin{subfigure}[t]{0.5\textwidth}
    \centering
    \caption{(a) Diamond of Gaussians}
    \includegraphics[width=0.95\linewidth]{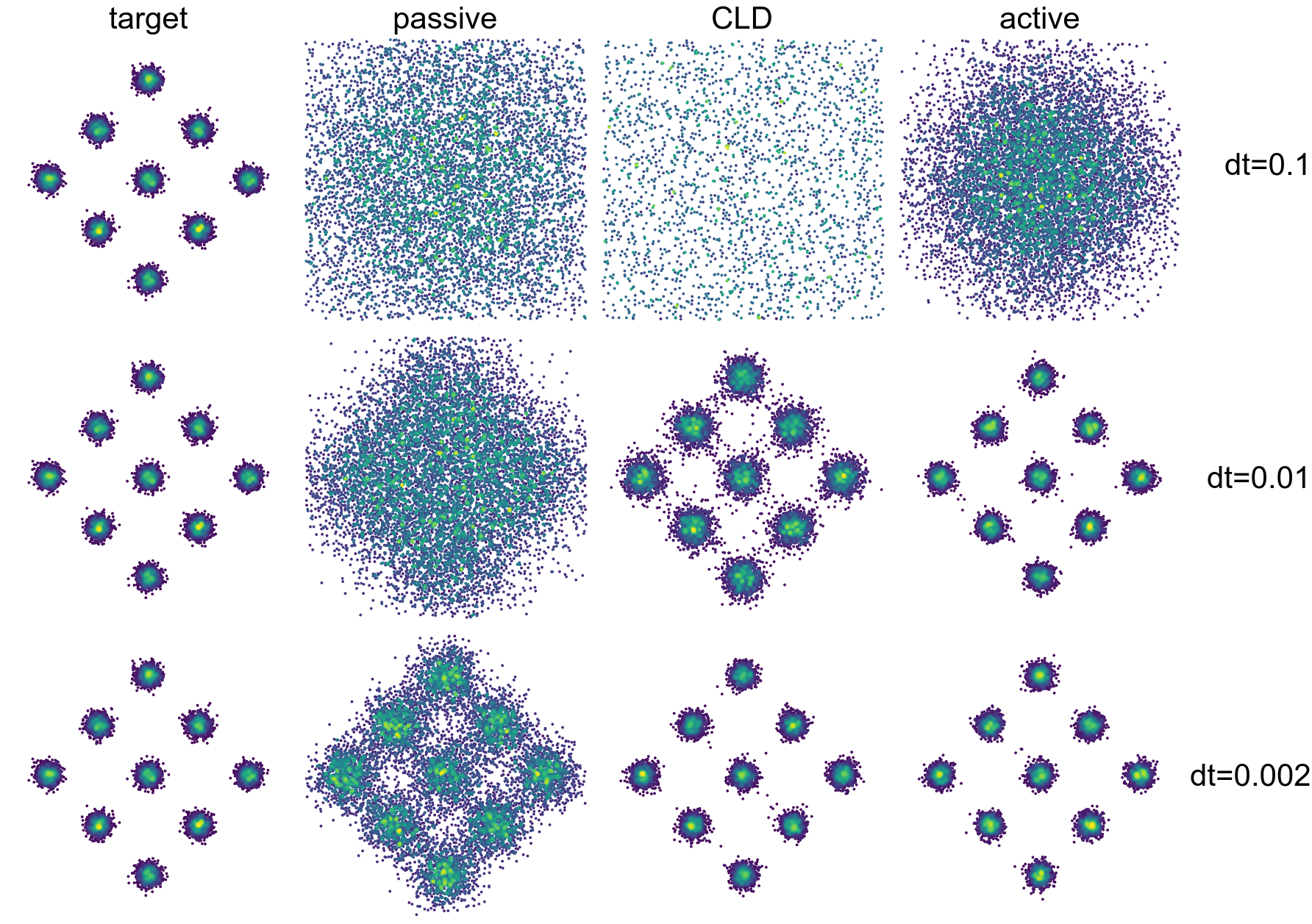}
    \label{si:fig:multigaussian:far}
    \end{subfigure}%
~
    \begin{subfigure}[t]{0.5\textwidth}
    \centering
    \caption{(b) Diamond of Gaussians with closer spacings between peaks}
    \includegraphics[width=0.95\linewidth]{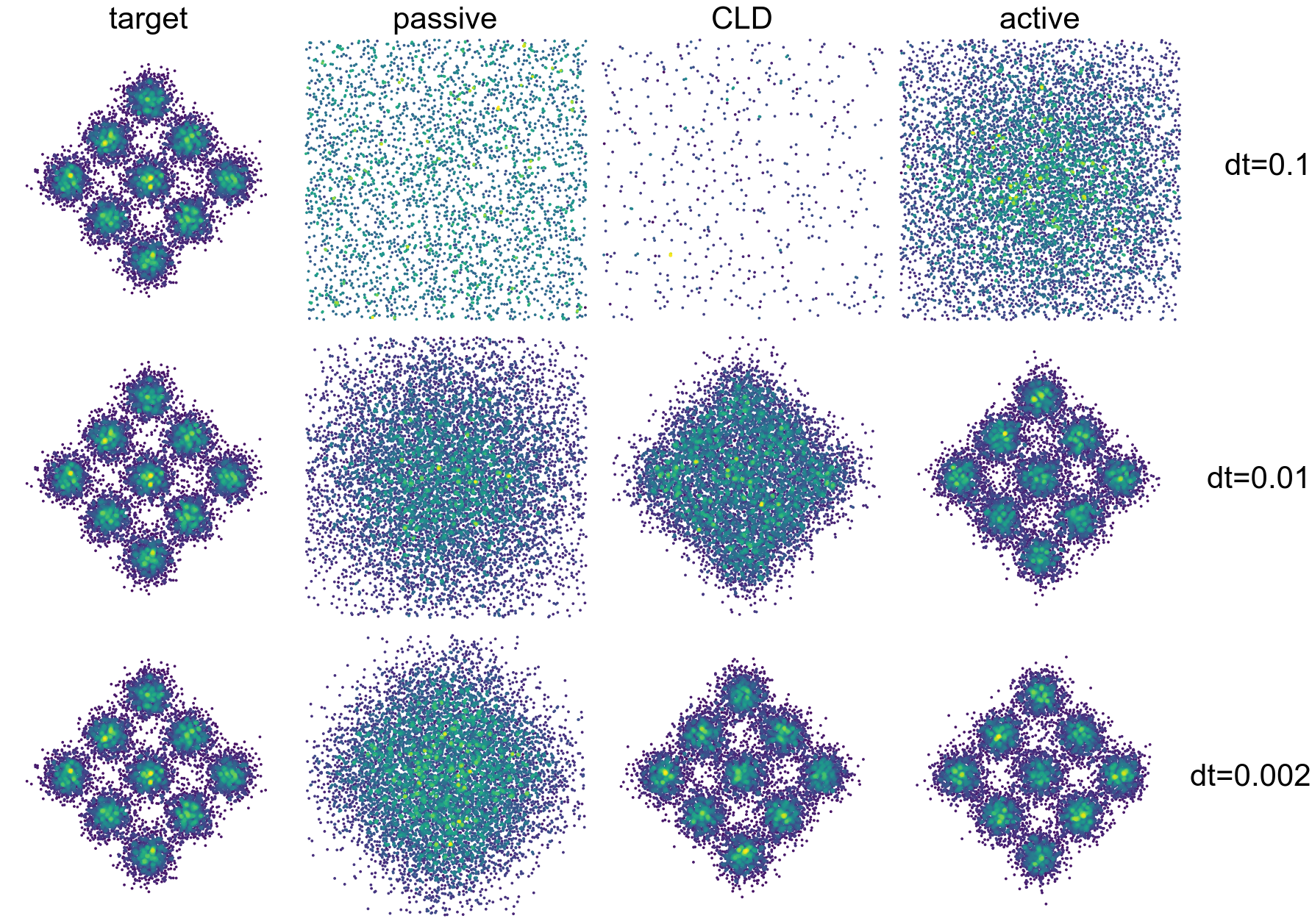}
    \label{si:fig:multigaussian:close}
    \end{subfigure}
~
    \caption{ Performance of various models with the score function approximated by a a neural network for Gaussian mixture target distributions at two peak separation distances ((a) $r=1/\sqrt{2}$ and (b) $r=\sqrt{2}/5$).}
    \label{si:fig:multigaussian}
\end{figure*}

Fig.~\ref{si:fig:swissrolls} examines the Swiss roll distributions. For large $dt$ ($=0.1$) all the methods, passive, active and CLD fail to resolve the coarse (locations of the swiss rolls) and finer details (the swiss roll spirals) of the distribution. At $dt=0.01$, CLD and active diffusion both resolve the position of the Swiss rolls, and active diffusion begins to capture the spiral features. At $dt=0.002$, CLD captures the spiral features of the Swiss rolls, while passive diffusion only shows faint traces of spiral structure (as evidenced in the point density indicated by green points in Fig.~\ref{si:fig:swissrolls:single}).

\begin{figure*}[htb]
    \begin{subfigure}[t]{0.5\textwidth}
    \centering
    \caption{(a) Swiss roll}
    \label{si:fig:swissrolls:single}
    \includegraphics[width=0.95\linewidth]{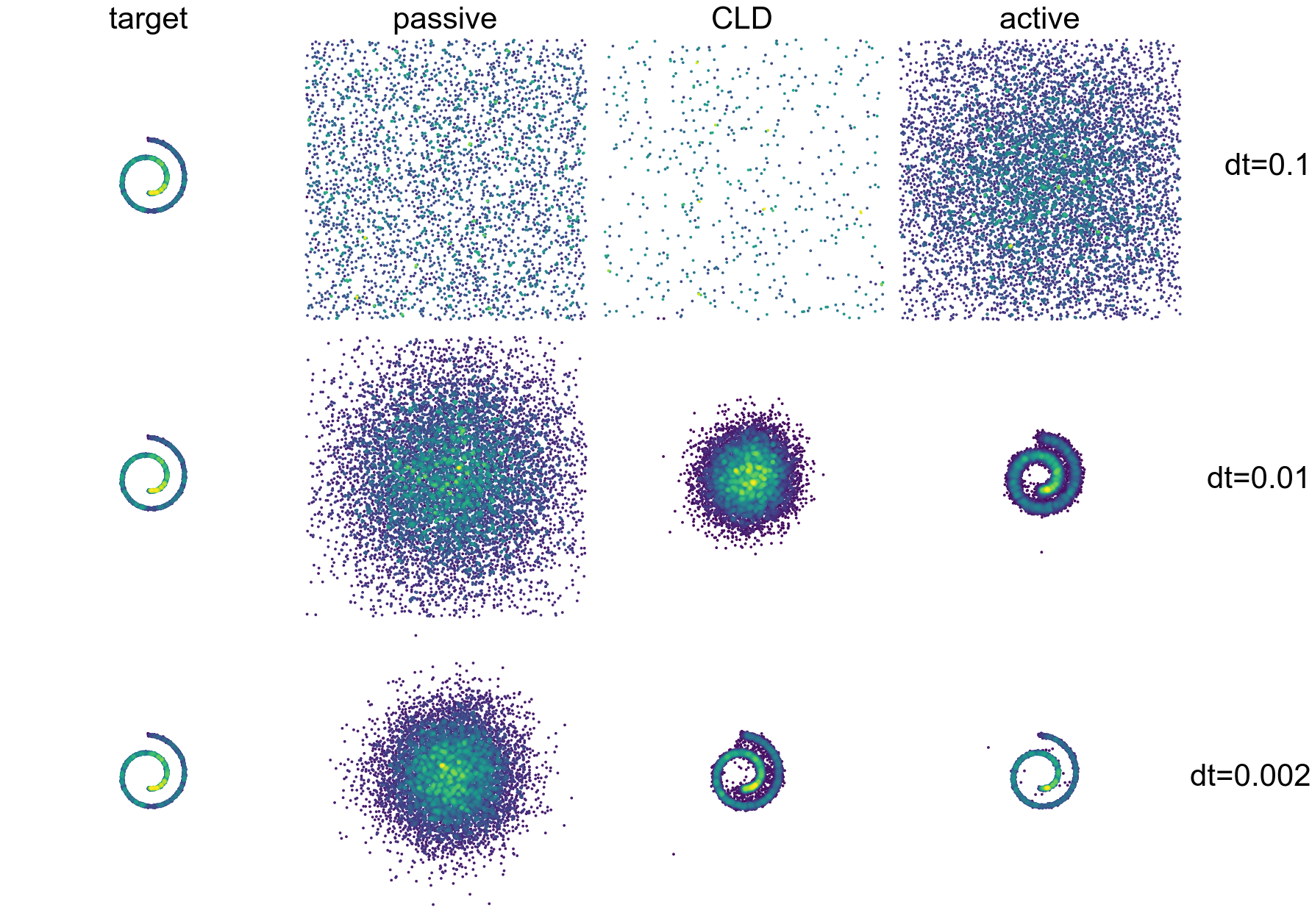}
    \end{subfigure}%
~
    \begin{subfigure}[t]{0.5\textwidth}
    \centering
    \caption{(b) Multiple Swiss rolls}
    \label{si:fig:swissrolls:multiple}
    \includegraphics[width=0.95\linewidth]{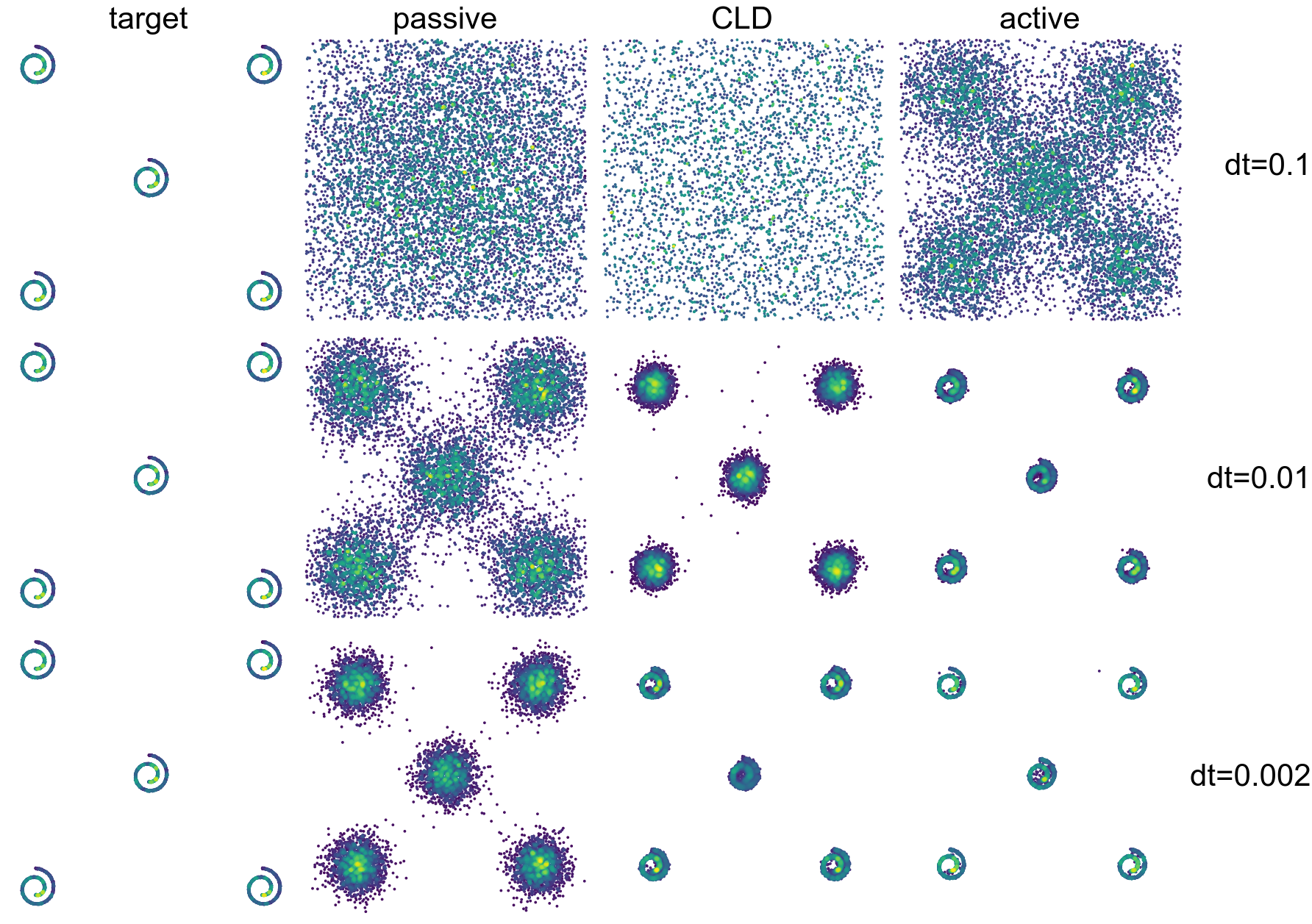}
    \end{subfigure}
~
    \caption{Performance of various models with the score function approximated by a a neural network for distributions consisting of a single (a) and multiple (b) Swiss rolls.}
    \label{si:fig:swissrolls}
\end{figure*}

\begin{figure*}[htb]
    \begin{subfigure}[t]{0.5\textwidth}
    \centering
    \caption{(a) Effect of $dt$ of the reverse diffusion process}
    \includegraphics[width=0.95\linewidth]{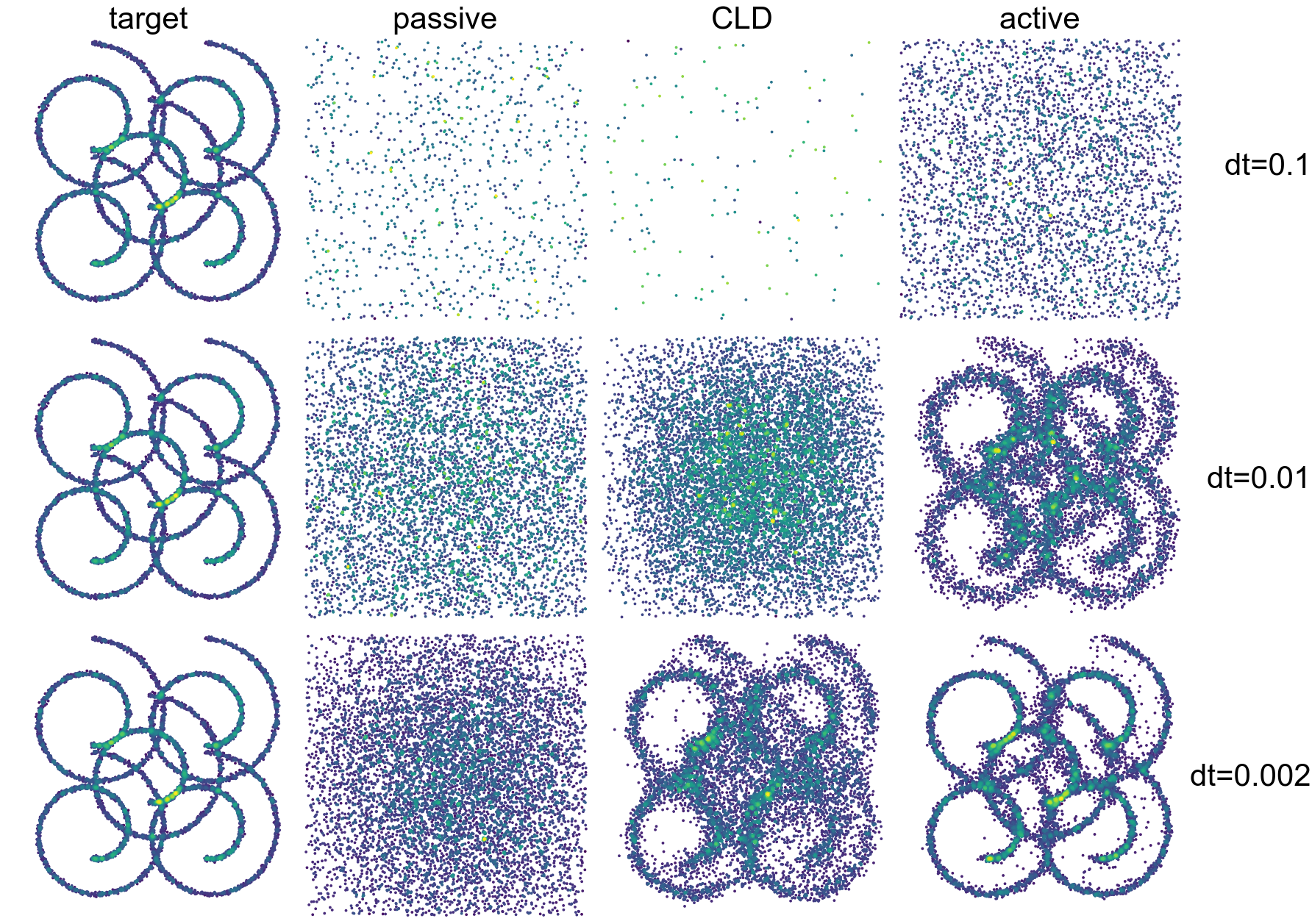}
    \label{si:fig:multimodal_swissroll_overlap:dt}
    \end{subfigure}%
~
    \begin{subfigure}[t]{0.5\textwidth}
    \centering
    \caption{(b) Effect of number of training iterations}
    \includegraphics[width=0.95\linewidth]{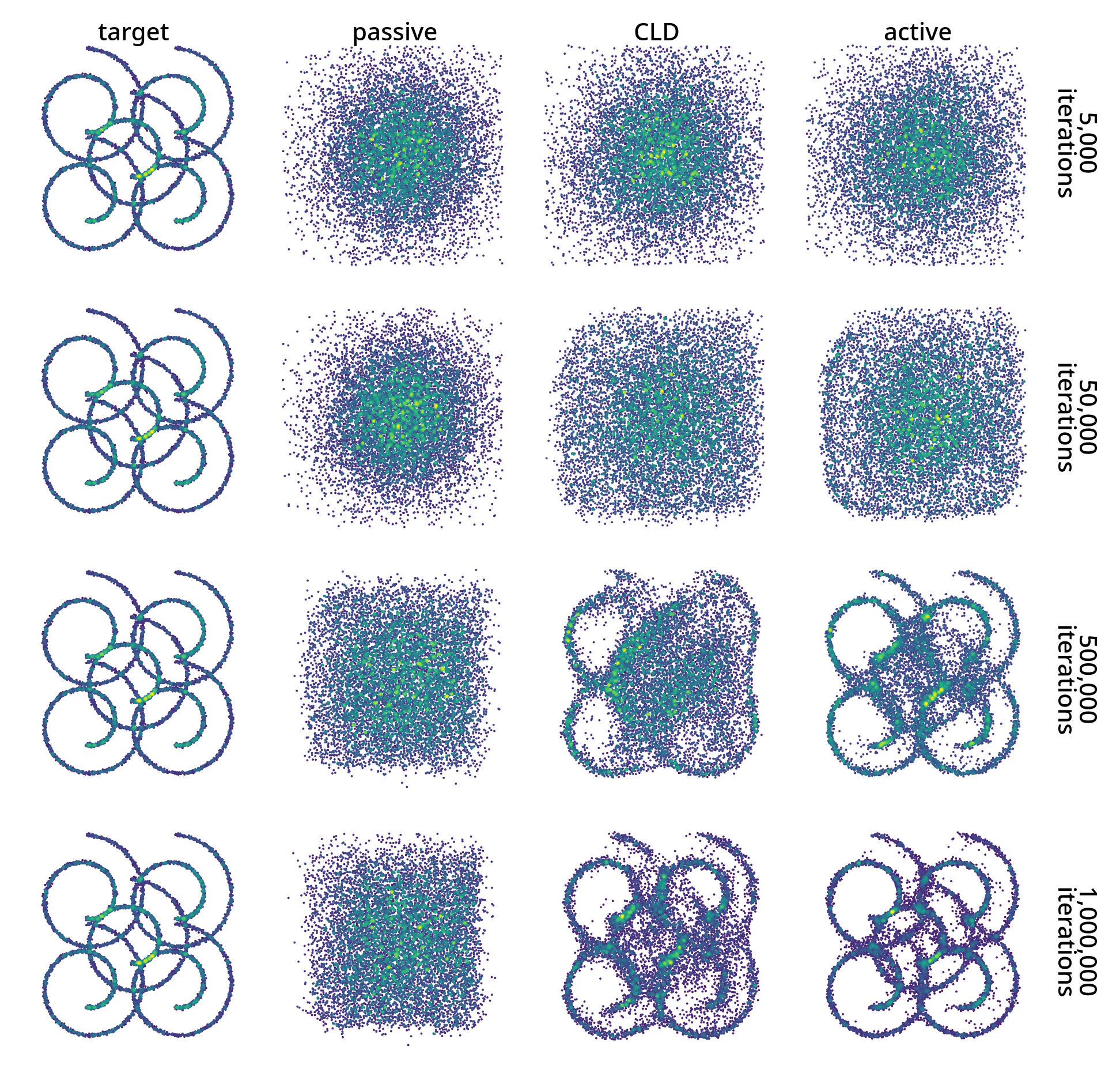}
    \label{si:fig:multimodal_swissroll_overlap:iters}
    \end{subfigure}
    \caption{Performance of various models with the score function approximated by a a neural network for distributions consisting of five overlapping Swiss rolls. (a) Effect of $dt$ on generated samples. (b) Effect of number of training iterations on generated samples. }
    \label{si:fig:multimodal_swissroll_overlap}
\end{figure*}

\FloatBarrier

\section{Alanine dipeptide training data generation}
\label{appssec:ADTrainingData}

Using GROMACS v2022.4~\cite{Berendsen1995GROMACS,Abraham2015GROMACS}, we prepared the alanine dipeptide system using the amber03 force field~\cite{Duan2003AMBER}, with explicit solvation using TIP3P~\cite{Jorgensen1983TIP3P} water molecules in a cubic box with length 1.2 nanometers with periodic boundary conditions in all directions. We also added NaCl at 0.15 molar concentration to represent physiological conditions. All electrostatics were treated using the Particle-Mesh-Ewald (PME) method~\cite{Darden1993PME} in GROMACS. We energy minimized the system for 50,000 steps using the steepest descent algorithm. After energy minimization, we held the alanine dipeptide position fixed and equilibrated the system under constant number, volume, and temperature (NVT) ensemble for 10 nanoseconds. We used the modified Berendsen thermostat (velocity-rescale) to control the temperature.~\cite{Bussi2007Vrescale} We then equilibrated the system further under the constant number, pressure, and temperature (NPT) ensemble for another 10 nanoseconds. Here, we maintained the temperature using the modified Berendsen thermostat and maintained the pressure using the Parrinello-Rahman barostat.~\cite{Parrinello1981Barostat} Pressure was maintained isotropically in X, Y, and Z directions. For both NVT and NPT equilibration, we maintained a temperature of 300 K. For NPT equilibration, we maintained a pressure of 1 bar, using a 2 ps time constant for the Parrinello-Rahman barostat. In all cases, we used the leap-frog molecular dynamics integrator with a 2 fs timestep within GROMACS. 

Using the resulting structure of the equilibrated system, we ran 1 microsecond of unbiased, brute force NPT dynamics on the alanine dipeptide, controlling temperature and pressure with the modified Berendsen thermostat and Parrinello-Rahman barostats respectively. To control the size of the dataset, we extract the conformation of the alanine dipeptide at every picosecond, leading to 1,000,000 conformations for 1 microsecond of simulation.

\section{Alanine Dipeptide Diffusion}
\label{appsec:ala_diff}

\begin{figure}[htb]
    \centering
    \includegraphics[width=0.75\linewidth]{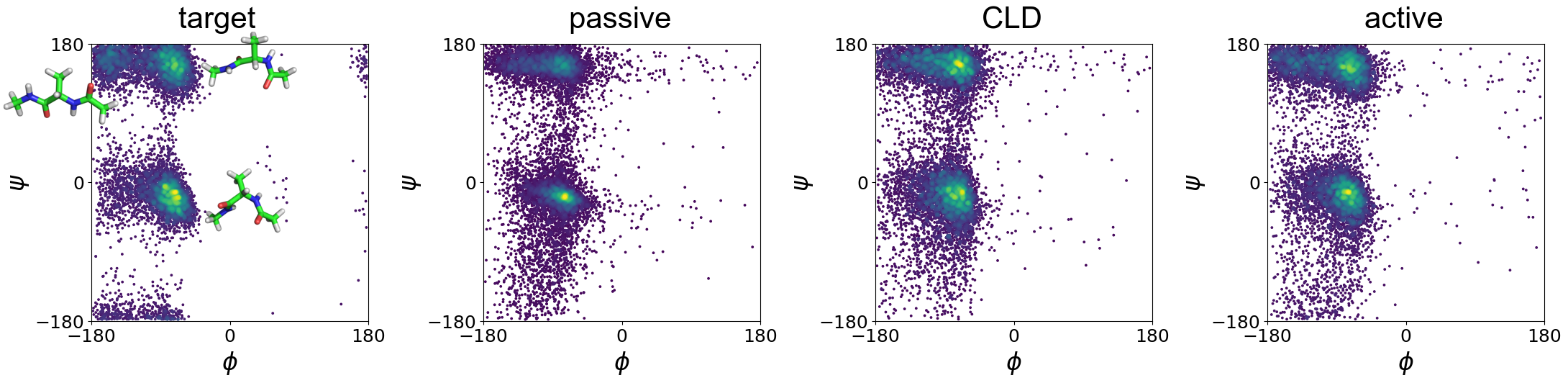}
    \caption{Ramachandran plots ($\phi$, $\psi$) in degrees for 1 $\mu$s of sampling for a water-solvated alanine dipeptide (left) and corresponding diffusion generated samples with passive (center left), CLD (center right), and active ($\tau=0.5$) (right).}
    \label{si:fig:alanine_dipeptide}
\end{figure}

The score model for the 2D distribution $(\phi, \psi)$ of alanine dipeptide was learned using a multi-layer perceptron with 4 hidden layers with 128 nodes each. The same architecture was used for the other 2D toy models. The 25D score model was learned using a U-net. The dataset was transformed into a $5\times5$ ``image'' to be used as input in the NN architecture used for image diffusion. 

Although the alanine dipeptide molecule has 28 unique parameters (bond lengths, bond angles, and dihedral angles), we reduce the dataset dimensionality by averaging some bond lengths. This reduction to 25 dimensions allowed us to treat the data like an image and use existing code functionality to demonstrate that active diffusion also outperforms passive diffusion for higher-dimensional toy datasets. As with the 2D alanine dipeptide dataset (Fig.~\ref{si:fig:alanine_dipeptide}), active diffusion performs better than passive diffusion or CLD at a lower number of iterations. 

\begin{figure*}[htb]
    \begin{subfigure}[t]{0.5\textwidth}
    \centering
    \caption{(a) First two components of PCA}
    \includegraphics[width=0.95\linewidth]{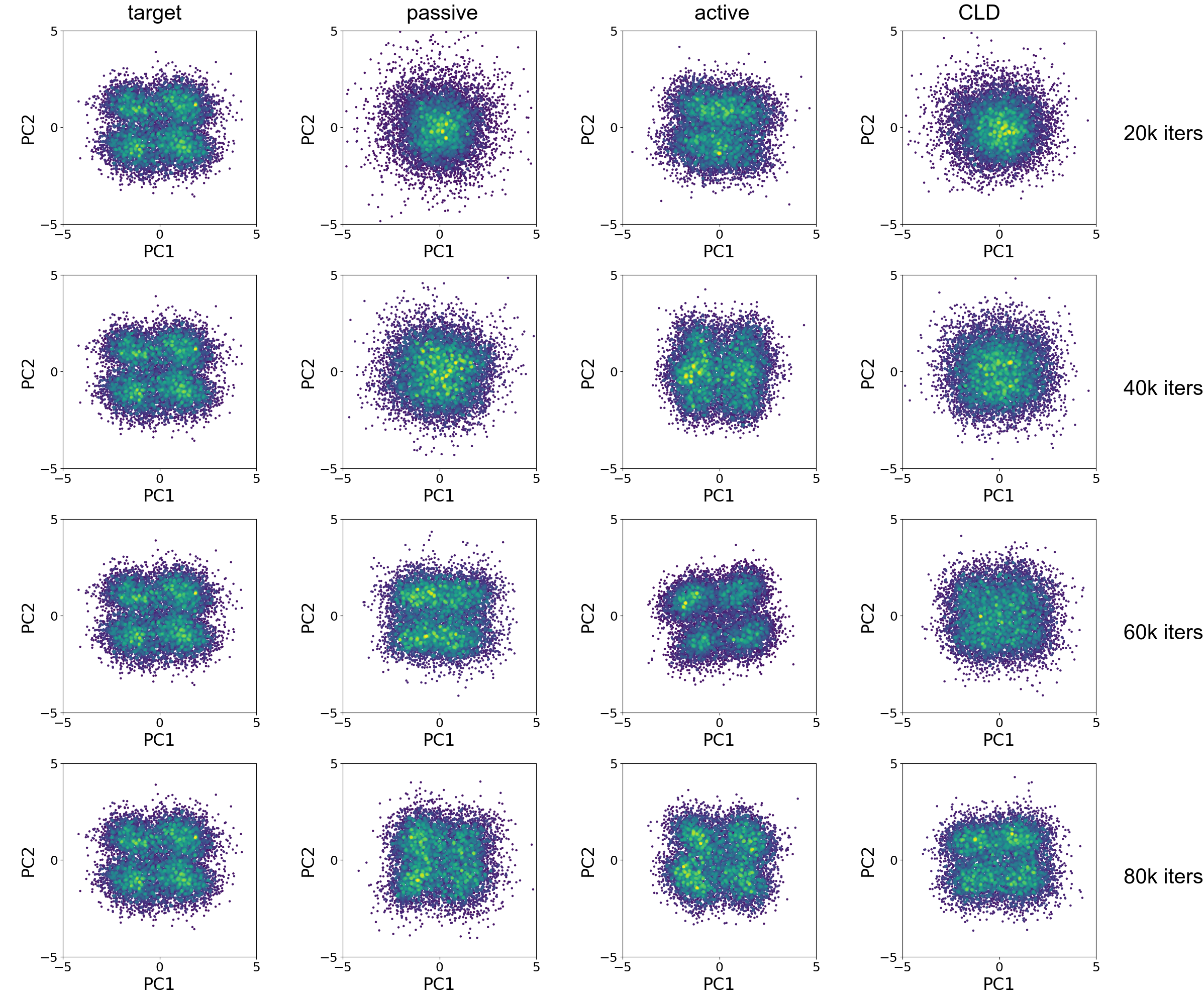}
    \label{si:fig:ala_25:pca}
    \end{subfigure}%
~
    \begin{subfigure}[t]{0.5\textwidth}
    \centering
    \caption{(b) $(\phi, \psi)$, taken from 25D diffusion data}
    \includegraphics[width=1.02\linewidth]{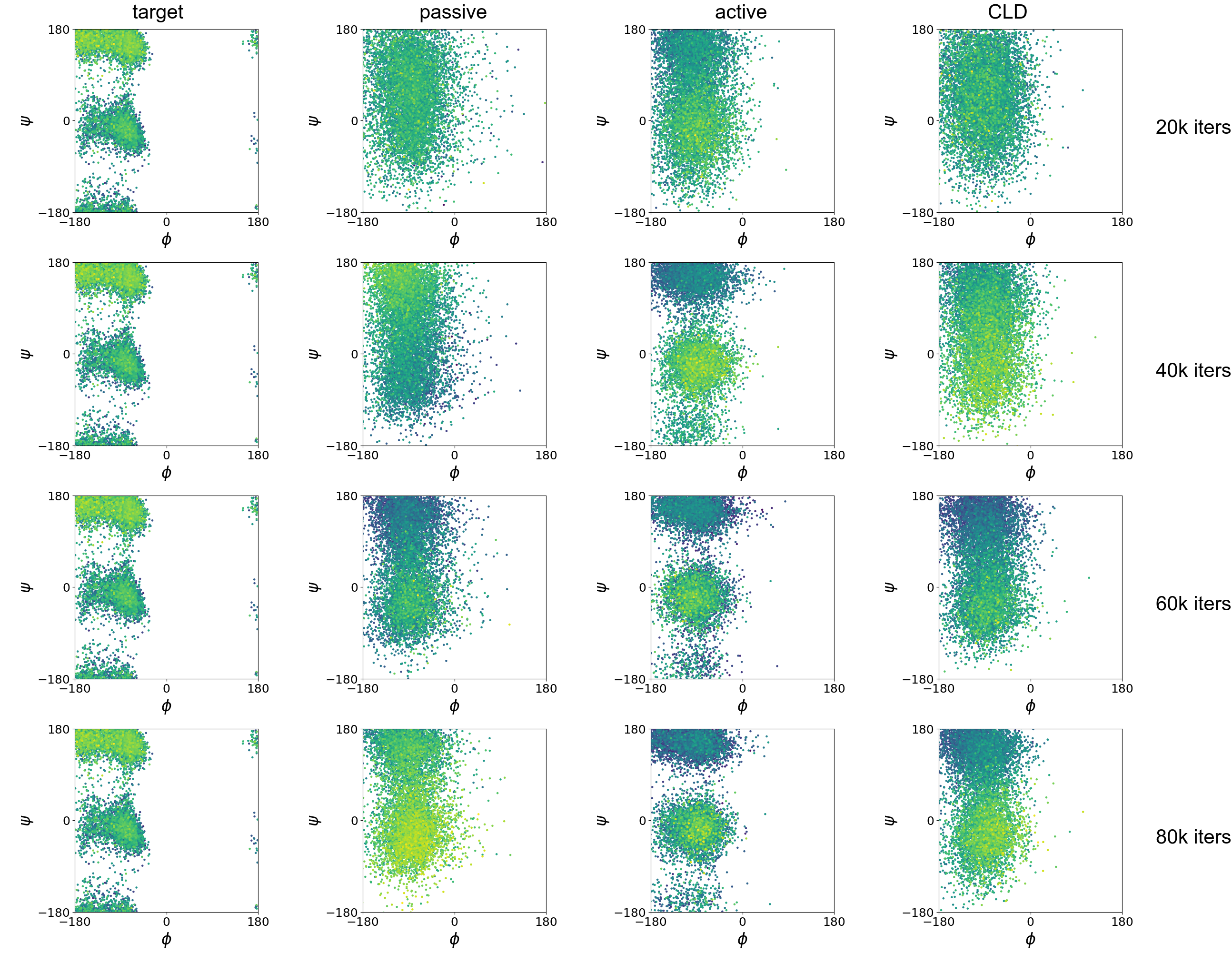}
    \label{si:fig:ala_25:dihedrals}
    \end{subfigure}%
    \caption{Effect of increasing number of training iterations on performance of 25D diffusion. (a) The first two principal components of training data (left column) and diffusion-generated data. Active diffusion PCA plot begins to resemble the training data at a lower number of training iterations. (b) Plot of dihedral angles generated by 25D diffusion. Active plot begins to resemble training data faster than passive diffusion or CLD.}
    \label{si:fig:ala_25}
\end{figure*}

\section{Speciation with active noise}
\label{sec:Speciation}
From Eq.~\eqref{eq:ActiveConditionalP} we integrate the $\bm{\eta}$ and $\bm{\eta}_0$ degrees, obtaining
\begin{align}
    P(\textbf{x};t) =& \int P(\textbf{x}_0) P(\textbf{x}|\textbf{x}_0;t) \\
    =& \int P(\textbf{x}_0) \exp(-\frac{(\textbf{x}-\textbf{x}_0 e^{-t})^2}{2\Delta_t^a}) \\
    =& \exp(-\frac{\textbf{x}^2}{2\Delta_t^a} + g(\textbf{x}))
\end{align}
where $\textbf{x}_0$ is the data at $t=0$, and $g(\textbf{x})$ and $\Delta_t^a$ are defined as
\begin{align}
    g(\textbf{x}) =& \log \left[ \int D\textbf{x}_0 P(\textbf{x}_0) \exp(-\frac{\textbf{x}_0^2 e^{-2t}}{2\Delta_t^a}-\frac{e^{-t}\textbf{x} \cdot \textbf{x}_0}{\Delta_t^a}) \right] \\
    \Delta_t^a =& \frac{T_a}{1+\tau} + \frac{\tau}{1-\tau} \left[ \frac{2\tau e^{-(1+\frac{1}{\tau})t}}{1+\tau} - e^{-2t} \right]
\end{align}

In the limit of large time, one can expand $g(\textbf{x})$ in terms of correlation functions. This yields
\begin{align}
    g(\textbf{x}) =& \frac{e^{-t}}{\Delta_t^a} \sum_{i=1}^d x_i \braket{x_{0,i}} + \frac{e^{-2t}}{2 \Delta_t^a}\sum_{i,j=1}^d x_i x_j [\braket{x_{0,i} x_{0,j}}  \nonumber \\
    &- \braket{x_{0,i}} \braket{x_{0,j}}] + \order{(xe^{-t})^3} \\
    \rm where & \ \langle \cdot \rangle = \mathbb{E}_{P(\mathbf{x_0})\exp(-\frac{\mathbf{x_0}^2e^{-2t}}{2\Delta_t^a})} [\cdot] \\
    \log(P_t(\textbf{x})) =& C + \frac{e^{-t}}{\Delta_t^a} \sum_{i=1}^d x_i \braket{x_{0,i}} \nonumber \\
    &- \frac{1}{2\Delta_t^a} \sum_{i,j=1}^d x_i M_{ij} x_j + \order{(\textbf{x}e^{-t})^3} \\
    M_{ij} =& \delta_{ij} - \frac{e^{-2t}}{\Delta_t^a}[\braket{x_{0,i} x_{0,j}} - \braket{x_{0,i}} \braket{x_{0,j}}]
\end{align}

The speciation time is given by the time when the curvature of $\log(P_t)$ changes shape. The matrix $M$ provides a quadratic form which helps $\log(P_t)$ change shape. The time for speciation, $t_s^a$, is thus the time when the largest eigenvalue of $M$ crosses $0$. At large times, the correlations in $M$, i.e., $\braket{x_{0,i} x_{0,j}} - \braket{x_{0,i}} \braket{x_{0,j}}$, can be substituted with the true covariance matrix of the target distribution, $C_0$. This leads to, 
\begin{align}
    e^{-2t_s^a} \cdot \rm \max_{\lambda} (C_0) =& \Delta_{t\to \infty}^a \\
    t_s^a =& \frac{1}{2} \log(\frac{\rm \max_{\lambda}(C_0) (1+\tau)}{T_a}) \label{eq:speciation_active}
\end{align}
where the notation $\rm max_{\lambda}(\cdot)$ denotes the maximum eigenvalue of $(\cdot)$.

\section{Belief Propagation on hierarchical data model}
\label{sec:bp_hierarchy}

In this section we provide details about the hierarchical model and belief propagation algorithm on the graph. We closely follow Ref.~\cite{sclocchi2025phase} and modify the analysis for our specific need. Natural images generally show a hierarchical structure with basic low level features like linea, patches, gradients etc being composed to form high level features which ultimately lead to the final image. To mimic this structure, a random hierarchical model (RHM) is developed. 

\subsection{Generation of Data}
The RHM defines a tree-like generative model with the following parameters:
\begin{enumerate}
    \item $L$ - Number of levels of the tree
    \item $s$ - Branching factor i.e. the number of children of every parent node
    \item $v$ - the alphabet size (each latent variable can take one of the $v$ values in $\{0,1,...,v-1 \}$)
    \item $m$ - the number of production rules per symbol
\end{enumerate}
Level $L$ is the root node and level $0$ are the leaf nodes which are the observations.
The production rules determine how data is generated. Each parent symbol $b \ (\in \{0,1,..,v-1\})$ at level $l$ can produce a certain combination of children. We encode this as a set $T^l_b$ of allowed tuples defined as,
\begin{align}
    T^l_b =& \{ \mathbf{t}^{(1)}, \mathbf{t}^{(2)}, .., \mathbf{t}^{(m)} \}, \ \rm m \ such \ tuples \\
    \mathbf{t}^{(i)} =& (t_1^{(i)}, t_2^{(i)}, .., t_s^{(i)}), \ t_k^{(i)} \in \{ 0,1,...,v-1 \} \forall k, \ k=\{1,2,..,s \}
\end{align}

The alphabets are represented as one-hot encoding in the $\mathbb{R}^v$ space. The dimensions of the observation (data) which corresponds to the leaf layer is thus given as, $\mathbb{R}^{v \times s^L}$. Data is generated in the following way:
\begin{enumerate}
    \item Sample the root node uniformly from the alphabets $\{0,1,...,v-1 \}$ \\
    \item For every layer, sample uniformly the production rules corresponding to the parent symbol and generate the child tuple
    \item iterate this till you reach the leaves
    \item represent the data as one-hot encoding of the alphabet at every index where the index ranges from $1$ to $s^L$
\end{enumerate}

\subsection{Optimal Denoising with Belief Propagation}
We use Bayes optimal denoising for the RHM. The data ($\vec{x}$) is first taken through a noising process, $\vec{x}(0) \to \vec{x}(t)$, then the probability, $P(\vec{x}(0)|\vec{x}(t))$ is computed exactly. Using this we compute the marginal probabilities of all the latent variables at all the layers using the message passing algorithm. Then we check if the marginals correctly predict the true labels at every layer of the data generation process.

\subsubsection{Belief Propagation}
In the case of RHM, the leaf nodes correspond to the input variables when messages(beliefs) are passed upwards and the root node is considered as the input variable when the messages are passed down. The factor nodes correspond to the production rules used for the creation of the data. During the upward pass, every parent collects the messages from its children and updates its beliefs and sends a message upward. This is iteratively performed till one reaches the root. Then in the downward phase, starting from the root, every node sends a downward message to its children and they update their beliefs and send subsequent messages. This is iteratively performed till one reaches the leaves. The beliefs during the upwards and downward phase are stored. The marginal is given as the product of the upward and downward beliefs.

Define $\mathbf{X}^{(0)}$ as the data vector and $\mathbf{X}^{(l)}$ be the high-level variable corresponding to the data at layer $\ell$. Let $\psi^{(\ell)}$ be any factor node connecting an $s$-tuple of low-level variables at layer $\ell - 1$, $\{\vec{X}_i^{(\ell-1)}\}_{i \in [s]}$, to a high-level variable $X_1^{(\ell)}$ at layer $\ell$. Without loss of generality, to lighten the notation, we rename the variables as $Y = X_1^{(\ell)}$, taking values $y \in \mathcal{A}$, and $X_i = X_i^{(\ell-1)}$, each taking values $x_i \in \mathcal{A}$. For each possible association $y \to x_1, \dots, x_s$, the factor node $\psi^{(\ell)}(y, x_1, \dots, x_s)$ takes values
\[
\psi^{(\ell)}(y, x_1, \dots, x_s) = 
\begin{cases} 
1, & \text{if } \{x_1,x_2...,x_s \} \in T_y^l \\ 
0, & \text{otherwise.} 
\end{cases}
\]

The BP upward and downward iterations for the (unnormalized) upward and downward messages respectively read
\begin{align*}
\tilde{v}_{\uparrow}^{(\ell+1)}(y) &= \sum_{x_1, \dots, x_s \in \mathcal{A}^{\otimes s}} \psi^{(\ell+1)}(y, x_1, \dots, x_s) \prod_{i=1}^s v_{\uparrow}^{(\ell)}(x_i), \\
\tilde{v}_{\downarrow}^{(\ell)}(x_1) &= \sum_{\substack{x_2, \dots, x_s \in \mathcal{A}^{\otimes (s-1)} \\ y \in \mathcal{A}}} \psi^{(\ell+1)}(y, x_1, \dots, x_s) \\
&\quad \times v_{\downarrow}^{(\ell+1)}(y) \prod_{i=2}^s v_{\uparrow}^{(\ell)}(x_i), \tag{5}
\end{align*}

where $v_{\rho}^{(\ell)}(x) = \frac{\tilde{v}_{\rho}^{(\ell)}(x)}{\sum_{x'} \tilde{v}_{\rho}^{(\ell)}(x')}, \rho \in \{\uparrow, \downarrow\}$. The downward iteration, reported for $x_1$, can be trivially extended to the other variables $x_i$ by permuting the position indices. The values of $v_{\uparrow}^{(0)}(x_i)$ and $v_{\downarrow}^{(L)}(y)$ are set by the initial conditions.

\subsubsection{Initialization of the leaf and root nodes}

For the root nodes, we initialize the downward messages, $v_{\downarrow}^{(L)}(y) = 1/v$, which corresponds to a uniform prior over the possible classes $\{0,1,...,v-1 \}$. 

For the leaf nodes, the initialization of the upward messages is a little involved. As defined previously, data is $\mathbf{X}^{(0)}$. The data is in fact a matrix of $s^L$ columns of vectors $\vec{X}^{(0)}_i \in \mathbb{R}^v$. Thus we can write without loss of generality $X_i^{(0)} = e_\gamma$, with $e_\gamma$ a canonical basis vector with $1$ in position $\gamma$ and $0$ everywhere else. Its continuous diffusion process takes place in $\mathbb{R}^v$: Given the value $X_i^{(0)} = x_i(t)$, we can compute the probability of its starting value $p(x_i(0)|x_i(t))$ using Bayes formula. This computation is performed independently for each input variable $i$, and therefore does not take into account the spatial correlations given by the generative model. The probabilities of Eq. 4 are used to initialize the BP upward messages $v_{\uparrow}^{(0)}(x_i) = p(x_i(0)|x_i(t))$ at the input variables. In our active case, we have an additional $\mathbf{\eta}^{(0)}$ associated with the data which leads to 
the computation of $p(x_i(0)|x_i(t),\eta_i(t))$ instead of $p(x_i(0)|x_i(t))$. From the passive and active processes defined in Eq.~\ref{eq:PassiveForward}, \ref{eq:Activex}, \ref{eq:ActiveEta}, \ref{eq:ActiveConditionalP}, one can easily compute these conditional probabilities after some messy algebra,
\begin{align}
    \text{Passive:  } p(x(0) =& e_\mu | x(t)) = \frac{1}{Z} \exp\left(\frac{e^{-t}}{T(1-e^{-2t})} x_\mu(t) \right) \\
    Z =& \sum_{\lambda} p(x(0)=e_{\lambda}|x(t)) \\
    \text{Active:  } p(x(0)=e_{\mu}|x(t),\eta(t))=& \frac{1}{Z} \exp\left( -\frac{(m_{11} \alpha -m_{12} e^{-t/\tau})e^{-kt} [(m_{11} \alpha -m_{12} e^{-t/\tau}) x_{\mu}(t) + (m_{22}e^{-t/\tau} - m_{12} \alpha) \eta_{\mu}(t)]}{ \Delta [(\tau/T_a) \Delta + m_{11}\alpha^2 - 2m_{12} e^{-t/\tau} \alpha + m_{22} e^{-2t/\tau}]} \right. \\
    & \left.+ \frac{e^{-kt}(m_{11} x_{\mu}(t) - m_{12}\eta_{\mu}(t))}{\Delta} \right)
\end{align}
where $m_{11}, m_{12}, m_{22}$ are as defined in Eq.~\ref{eq:m11}, \ref{eq:m12}, \ref{eq:m22}, $\alpha = \frac{e^{-t/\tau} - e^{-kt}}{k-(1/\tau)}$, $\Delta = m_{11}m_{22} - m_{12}^2$.

We run three different simulations, one with passive noising process, one with active noising process and a third set with a passive temperature ``equivalent'' to the active temperature. The equivalence is set by the variance of the data dimension at the end of the nosiing process. For passive noise it is given by the passive temperature, $T$. For the active nosiing process, the final variance on the data dimension is given by, $T_a/(k(1+k\tau))$. The third set of simulation is carried out using a passive noising process with $T=T_a/(k(1+k\tau))$.
\end{widetext}

\section{Cifar-10 diffusion}
\label{app:Cifar}
We use a DDPM/NCSN++-style U-Net backbone on $32{\times}32$ CIFAR-10 images. The base width is 128 with channel multipliers $(1,2,2,2)$, yielding feature dims $\{128,256,256,256\}$ across four resolution levels, each with 4 residual blocks. Timesteps are embedded with sinusoidal features (dim 256) followed by an MLP (4$\times$ expansion) and injected into each residual block. Each block uses GroupNorm, SiLU, and two $3{\times}3$ convolutions (with a $1{\times}1$ skip when needed). Downsampling uses a $3{\times}3$ stride-2 conv; upsampling uses nearest-neighbor interpolation followed by a $3{\times}3$ conv, with skip concatenation. Self-attention is applied at $16{\times}16$ resolution, and the bottleneck is ResBlock–Attn–ResBlock. The output head is GroupNorm+SiLU+$3{\times}3$ conv. The passive model uses 3 input/output channels (RGB). The active model uses 6 input/output channels (RGB concatenated with 3 $\eta$ channels). 

\section{Code}
The code for active diffusion on MNIST dataset can be found at, \href{https://github.com/EvilBwala/Active_Diffusion_MNIST}{Active Diffusion MNIST} and the code for Belief Propagation on the Random Hierarchical model can be found at, \href{https://github.com/EvilBwala/BP_RHM_Active}{BP RHM Active}.

\FloatBarrier

\end{document}